\documentclass[twoside,11pt]{article}

\usepackage{blindtext}

\usepackage{jmlr2e}

\usepackage[utf8]{inputenc} % allow utf-8 input
\usepackage[T1]{fontenc}    % use 8-bit T1 fonts
\usepackage{url}            % simple URL typesetting
\usepackage{booktabs}       % professional-quality tables
\usepackage{amsfonts}       % blackboard math symbols
\usepackage{nicefrac}       % compact symbols for 1/2, etc.
\usepackage{microtype}      % microtypography
\usepackage[table]{xcolor}         % colors

\usepackage[acronym]{glossaries}
\usepackage{graphicx}
\usepackage{placeins}
\usepackage{listings}
\usepackage[framemethod=TikZ]{mdframed}
\usepackage{tikz}
\usetikzlibrary{shadows,calc}
\usepackage[T1]{fontenc}
\usepackage{comment}
\usepackage{colortbl}
\usepackage{mdframed}
\usepackage{adjustbox}
\usepackage{changepage}
\usepackage[group-separator={,}]{siunitx}
\usepackage{enumitem}
\usepackage{wrapfig}
\setlist[itemize]{leftmargin=0.5cm}
\usepackage{cleveref}
\usepackage{hyperref}
\hypersetup{
    hidelinks,
    pdfauthor={Jonas Eschmann, Dario Albani, Giuseppe Loianno},
    pdftitle={RLtools: A Fast, Portable Deep Reinforcement Learning Library for Continuous Control},
    pdfsubject={Deep Reinforcement Learning (RL) can yield capable agents and control policies in several domains but is commonly plagued by prohibitively long training times. Additionally, in the case of continuous control problems, the applicability of learned policies on real-world embedded devices is limited due to the lack of real-time guarantees and portability of existing libraries. To address these challenges, we present RLtools, a dependency-free, header-only, pure C++ library for deep supervised and reinforcement learning. Its novel architecture allows RLtools to be used on a wide variety of platforms, from HPC clusters over workstations and laptops to smartphones, smartwatches, and microcontrollers. Specifically, due to the tight integration of the RL algorithms with simulation environments, RLtools can solve popular RL problems up to 76 times faster than other popular RL frameworks. We also benchmark the inference on a diverse set of microcontrollers and show that in most cases our optimized implementation is by far the fastest. Finally, RLtools enables the first-ever demonstration of training a deep RL algorithm directly on a microcontroller, giving rise to the field of Tiny Reinforcement Learning (TinyRL). The source code as well as documentation and live demos are available through our project page at https://rl.tools.},
    pdfkeywords={Reinforcement Learning, Continuous Control, Deep Learning, TinyRL}
}

\definecolor{primary_color}{HTML}{7DB9B6}
\definecolor{primary_color_readable}{HTML}{639694}

\newcommand{\rltools}{\textcolor{primary_color_readable}{\textbf{RLtools}}}
\newcommand{\grayrow}{\rowcolor{gray!10}}

\newacronym{mdp}{MDP}{Markov Decision Process}
\newacronym{rl}{RL}{Reinforcement Learning}
\newacronym{dl}{DL}{Deep Learning}
\newacronym{mav}{MAV}{Micro Aerial Vehicle}
\newacronym{ml}{ML}{Machine Learning}
\newacronym{dnn}{DNN}{Deep Neural Network}
\newacronym{trpo}{TRPO}{Trust Region Policy Optimization}
\newacronym{ddpg}{DDPG}{Deep Deterministic Policy Gradient}
\newacronym{gae}{GAE}{Generalized Advantage Estimation}
\newacronym{ppo}{PPO}{Proximal Policy Optimization}
\newacronym{gps}{GPS}{Guided Policy Search}
\newacronym{cnn}{CNN}{Convolutional Neural Network}
\newacronym{rnn}{RNN}{Recurrent Neural Network}
\newacronym{relu}{ReLU}{Rectified Linear Unit}
\newacronym{tqc}{TQC}{Trucated Quantile Critics}
\newacronym{hpc}{HPC}{High-Performance Computing}
\newacronym{fcu}{FCU}{Flight Controller Unit}
\newacronym{fma}{FMA}{Fused Multiply-Accumulate}
\newacronym{nvcc}{nvcc}{NVIDIA CUDA Compiler}
\newacronym{stl}{STL}{Standard Template Library}
\newacronym{td3}{TD3}{Twin Delayed Deep Deterministic policy gradient}
\newacronym{vmt}{VMT}{Virtual Method Table}
\newacronym{rpm}{RPM}{Revolutions Per Minute}
\newacronym{uav}{UAV}{Unmanned Aerial Vehicle}
\newacronym{rk4}{RK4}{$4^{\text{th}}$ order Runge-Kutta}
\newacronym{gemm}{GEMM}{GEneral Matrix Multiply}
\newacronym{sfinae}{SFINAE}{Substitution Failure Is Not An Error}
\newacronym{ram}{RAM}{Random Access Memory}
\newacronym{iot}{IoT}{Internet of Things}
\newacronym{gcc}{GCC}{GNU Compiler Collection}
\newacronym{dsp}{DSP}{Digital Signal Processor}
\newacronym{sac}{SAC}{Soft Actor-Critic}
\newacronym{kl}{KL}{Kullback-Leibler}
\newacronym{wasm}{WASM}{WebAssembly}
\newacronym{ui}{UI}{User Interface}
\newacronym{amx}{AMX}{Apple Matrix Coprocessor}
\newacronym{tinyrl}{TinyRL}{Tiny Reinforcement Learning}
\newacronym{tinyml}{TinyML}{Tiny Machine Learning}
\newacronym{ode}{ODE}{Ordinary Differential Equation}
\newacronym{pid}{PID}{Proportional–Integral–Derivative}
\newacronym{mlp}{MLP}{Multilayer Perceptron}
\newacronym{mpc}{MPC}{Model Predictive Control}
\newacronym{blas}{BLAS}{Basic Linear Algebra Subprograms}
\newacronym{simd}{SIMD}{Single Instruction, Multiple Data}
\newacronym{avx}{AVX}{Advanced Vector Extensions}
\newacronym{sse}{SSE}{Streaming SIMD Extensions}
\newacronym{iqm}{IQM}{Inter Quantile Mean}

\usepackage{lastpage}

\usepackage{lastpage}
\jmlrheading{25}{2024}{1-\pageref{LastPage}}{2/24; Revised
8/24}{9/24}{24-0248}{Jonas Eschmann, Dario Albani, and Giuseppe Loianno}
\ShortHeadings{\rltools{}: A Fast, Portable Deep Reinforcement Learning Library}{Eschmann, Albani, and Loianno}

\firstpageno{1}

\begin{document}

\title{\rltools{}: A Fast, Portable Deep Reinforcement Learning Library for Continuous Control}

\author{%
\begin{center}
\vspace{-7mm}
Jonas Eschmann$^{1,2}$ \quad Dario Albani$^{2}$ \quad Giuseppe Loianno$^1$ \\
$^1$\small{New York University} \quad $^2$\small{Technology Innovation Institute}\\
\texttt{\{jonas.eschmann,loiannog\}@nyu.edu}\\
\texttt{dario.albani@tii.ae} 
\end{center}
\vspace{-0.5cm}
}

\editor{Alexandre Gramfort}
\maketitle
\vspace{-1.2cm}
\begin{center}
\includegraphics[width=0.85\textwidth]{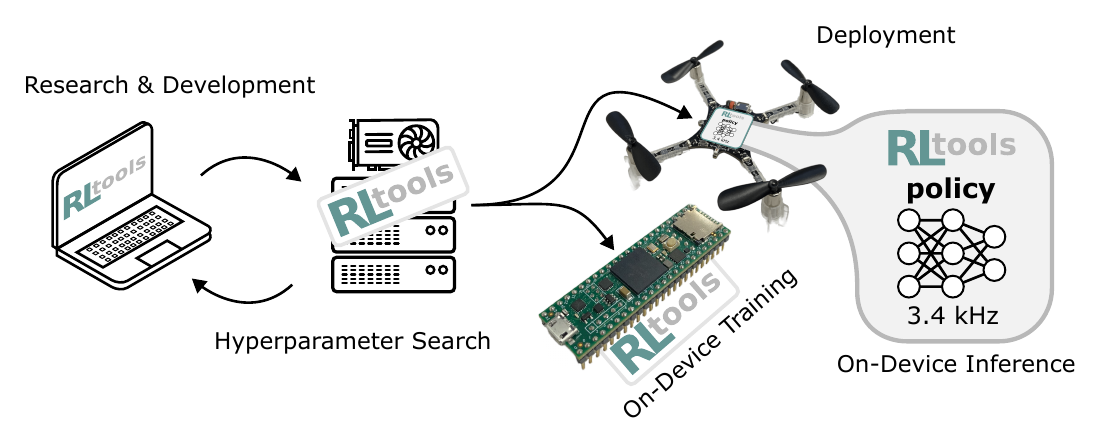}
\end{center}
\vspace{-5mm}
\begin{abstract}
Deep \gls*{rl} can yield capable agents and control policies in several domains but is commonly plagued by prohibitively long training times. Additionally, in the case of continuous control problems, the applicability of learned policies on real-world embedded devices is limited due to the lack of real-time guarantees and portability of existing libraries. 
To address these challenges, we present \rltools{}, a dependency-free, header-only, pure C++ library for deep supervised and reinforcement learning.
Its novel architecture allows \rltools{} to be used on a wide variety of platforms, from HPC clusters over workstations and laptops to smartphones, smartwatches, and microcontrollers. 
Specifically, due to the tight integration of the \gls*{rl} algorithms with simulation environments, \rltools{} can solve popular \gls*{rl} problems up to \num{76} times faster than other popular \gls*{rl} frameworks.
We also benchmark the inference on a diverse set of microcontrollers and show that in most cases our optimized implementation is by far the fastest. Finally, \rltools{} enables the first-ever demonstration of training a deep \gls*{rl} algorithm directly on a microcontroller, giving rise to the field of \gls*{tinyrl}. 
The source code as well as documentation and live demos are available through our project page at \href{https://rl.tools}{https://rl.tools}.
\end{abstract}

\vspace{2mm}
\begin{keywords}
  Reinforcement Learning, Continuous Control, Deep Learning, TinyRL
\end{keywords}

\section{Introduction}
\label{introduction}
Continuous control is a ubiquitous and pervasive problem in a diverse set of domains such as robotics, high-frequency decision-making in financial markets or the automation of chemical plants and smart grid infrastructure. 
Taking advantage of the recent progress in \gls*{dl} that is spilling over into decision-making in the form of \gls*{rl},
agents derived using deep \gls*{rl} have already attained impressive performance in a range of decision-making problems, like games and particularly continuous control. 
Despite these achievements, the real-world adoption of \gls*{rl} for continuous control is hindered by prohibitively long training times as well as a lack of support for the deployment of trained policies on real-world embedded devices. Long training times obstruct rapid iteration in the problem space (reward function design, hyperparameter tuning, etc.) while deployment on computationally severely limited embedded devices is necessary to control the bulk of physical systems such as: robots, automotive components, medical devices, smart grid infrastructure, etc. 
In non-physical systems, such as financial markets, the need for high-frequency decision-making leads to similar real-time requirements which cannot be fulfilled by current deep \gls*{rl} libraries.
Hence, to address these challenges we present \rltools{}, a dependency-free, header-only pure C++ library for deep supervised and reinforcement learning combining the following contributions:
\begin{itemize}[nosep]
\item \textbf{Novel Architecture}: We describe the innovations in the software design of the library which allow for unprecedented training and inference speeds on a wide variety of devices from \gls*{hpc} clusters over workstations and laptops to smartphones, smartwatches and microcontrollers.
\item \textbf{Implementation}: We contribute a modular, highly portable, and efficient implementation of the aforementioned architecture in the form of open-source code, documentation, and test cases.
\item \textbf{Fastest Training}: We demonstrate large speedups in terms of wall-clock training time.
\item \textbf{Fastest Inference}: We demonstrate large speedups in terms of the inference time of trained policies on a diverse set of common microcontrollers. 
\item \textbf{\gls*{tinyrl}}: By utilizing \rltools{}, we successfully demonstrate, the first-ever training of a deep \gls*{rl} algorithm for continuous control directly on a microcontroller. 
\end{itemize}

\FloatBarrier
\section{Related Work}
\label{sec:related_work}
% \textbf{Deep \gls*{rl} libraries and frameworks} \hspace{0.1cm}
Multiple deep \gls*{rl} frameworks and libraries have been proposed, many of which cover algorithmic research, with and without abstractions (Acme \citep{hoffman_acme_2022}, skrl \citep{serrano2023skrl} and CleanRL \citep{huang_cleanrl_2021} respectively). Other frameworks and libraries focus on comprehensiveness in terms of the number of algorithms included (RLlib \citep{liang_rllib_2018}, ReinforcementLearning.jl \citep{Tian2020Reinforcement}, MushroomRL \citep{deramo_mushroomrl_2020}, Stable-Baselines3 \citep{ran_stable-baselines3_2021}, ChainerRL \citep{fujita_chainerrl_2021}), Tianshou \citep{weng_tianshou_2022}, and TorchRL \citep{bou2023torchrl}.  
In contrast to these aforementioned solutions, \rltools{} aims at fast iteration in the problem space in the form of e.g., reward function design \citep{eschmann2021reward} and hyperparameter optimization. In the problem space, the algorithmic intricacies and variety of the algorithms matter less than the robustness, training speed, and final performance as well as our understanding of how to train them reliably. 
From the formerly mentioned \gls*{rl} frameworks and libraries RLlib \citep{liang_rllib_2018} is the most similar in terms of its mission statement being on quick iteration and deployment (cf. benchmark comparisons wrt. this goal in Section \ref{sec:results}). 
By focusing on iteration in the space of problems and subsequent deployment to real-time platforms, we also draw parallels between \rltools{}  and the ACADOS software \citep{verschueren_acadosmodular_2022} for synthesizing \glspl*{mpc} with \rltools{} aspiring to be its \gls*{rl} equivalent. 

\FloatBarrier
\section{Approach}
\label{sec:approach}

Taking the last handful of years of progress in \gls*{rl} for continuous control, it can be observed that the most prominent models used as function approximators are still relatively small, fully-connected neural networks. In Appendix \ref{appendix:analysis_drl} we analyze the architectures used in deep \gls*{rl} for continuous control and justify the focus of \rltools{} on (small) fully-connected neural networks.
Based on these observations, we conclude that the great flexibility provided by automatic differentiation frameworks like TensorFlow or PyTorch might not be necessary for applying \gls*{rl} to many continuous control problems. We believe that there is an advantage in trading-off the flexibility in the model architecture of the function approximators for the overall training speed. 
Reducing the training time and increasing the training efficiency saves energy, simplifies reproducibility and democratizes access to state of the art \gls*{rl} methods. Furthermore, fast training facilitates principled hyperparameter search which in turn improves comparability. 

\noindent\textbf{Architecture}
\label{sec:approach_architecture}
Our software architecture is guided by the previous observation and hence by maximizing the training time efficiency without sacrificing returns. Additionally, we want the software to be able to run across many different accelerators and devices (CPUs, GPUs, microcontrollers, and other accelerators) so that trained policies can also directly be deployed on microcontrollers and take advantage of device-specific instructions to run at high frequencies with hard realtime guarantees. This also entails that \rltools{} does not rely on any dependencies because they might not be available on the target microcontrollers. 

To attain maximum performance, we integrate the different components of our library as tightly as needed while maintaining as much flexibility and modularity as possible. To enable this goal, we heavily rely on the C++ templating system. Leveraging template meta-programming, we can provide the compiler with a maximum amount of information about the structure of the code, enabling it to be optimized heavily. We particularly make sure that the size of all loops is known at compile time such that the compiler can optimize them via inlining and loop-unrolling (cf. Appendix \ref{appendix:approach_programming_paradigm} and \ref{appendix:ablation_study}). 
Leveraging pure C++ without any dependencies, we implement the following major components: \textbf{Deep Learning} (MLP, backpropagation, Adam, etc.), \textbf{Reinforcement Learning} (GAE, PPO, TD3, SAC), and \textbf{Simulation} (Pendulum, Acrobot, Quadrotor, Racing Car, MuJoCo interface). 
We implement \rltools{} in a modular way by using a novel static multiple-dispatch paradigm inspired by (dynamic) multiple-dispatch which was popularized by the Julia programming language \cite{bezanson_julia_2012}. We highly recommend taking a look at the code example and explanation in Appendix \ref{appendix:approach_programming_paradigm} as well as the ablation study in Appendix \ref{appendix:ablation_study} measuring the impact of different components and optimizations.

\section{Results}
\label{sec:results}
\begin{figure}
  \centering
  \begin{minipage}{0.49\textwidth}
    \includegraphics[width=1.00\linewidth]{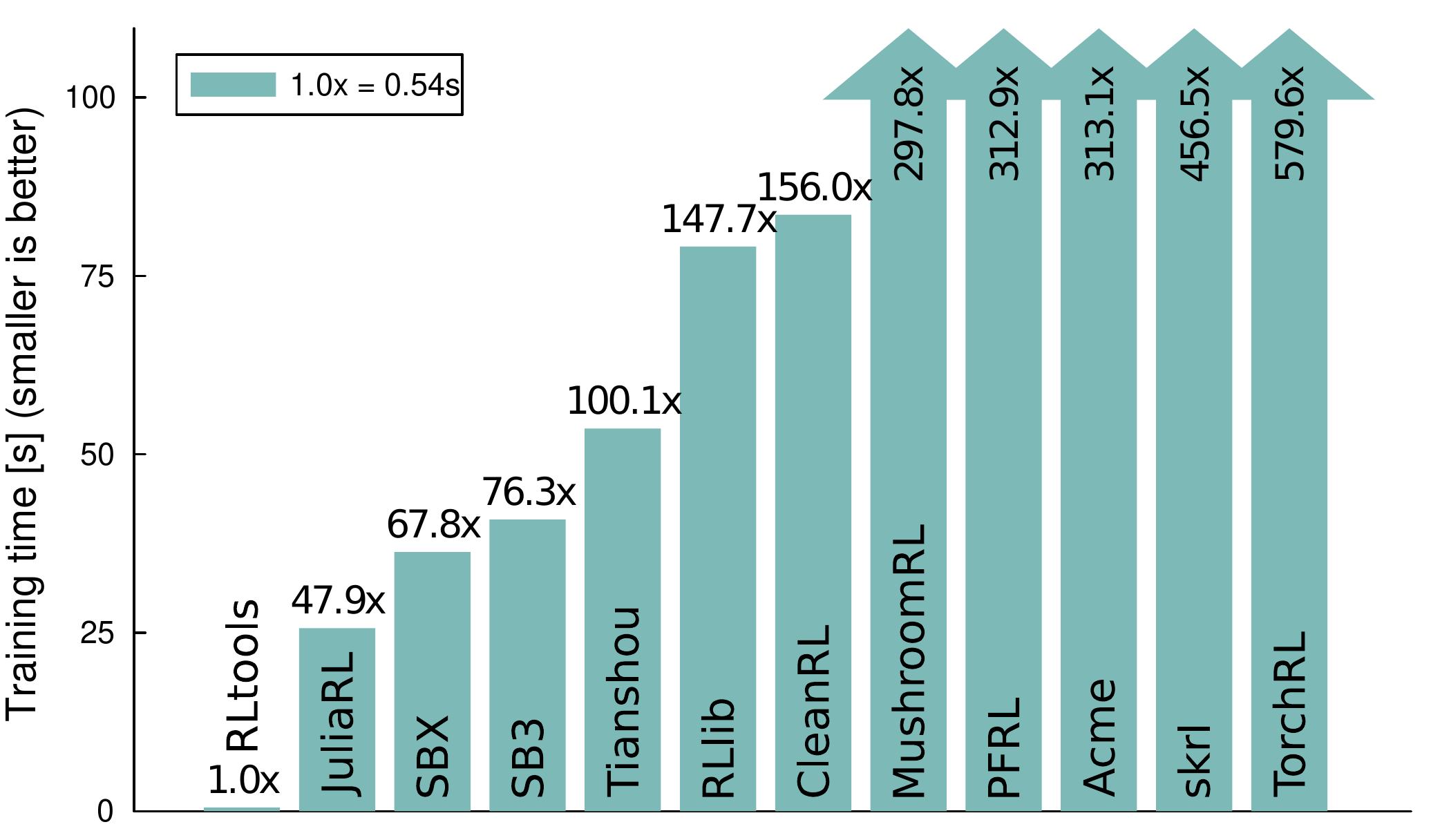}
    \centering
    \vspace{-10mm}
    \caption{\gls*{ppo}: \texttt{Pendulum-v1} ($300000$ steps)}
    \label{fig:benchmark_ppo_results_medium}
  \end{minipage}
  \begin{minipage}{0.49\textwidth}
    \includegraphics[width=1.00\linewidth]{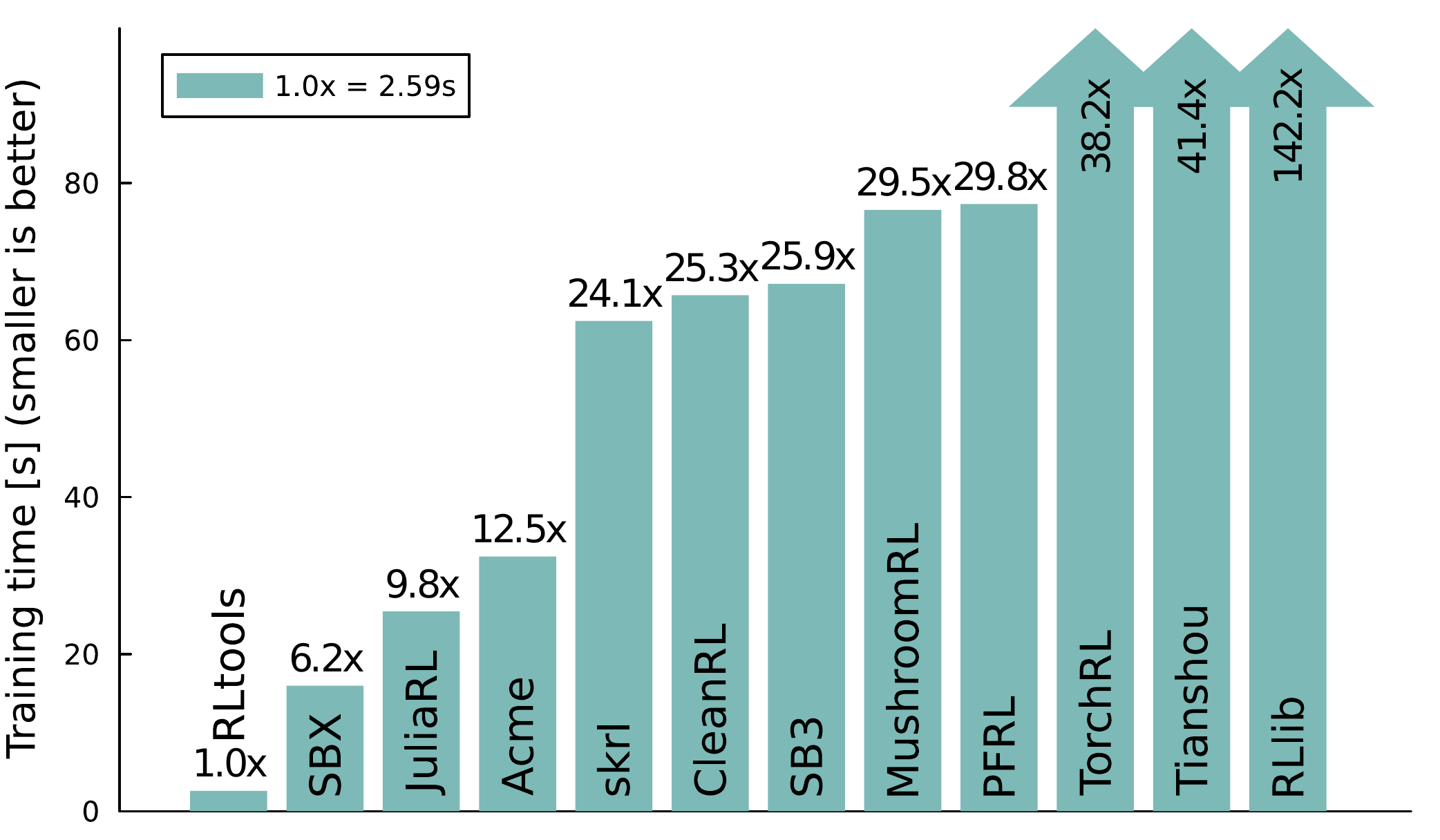}
    \centering
    \vspace{-10.5mm}
    \caption{\gls*{sac}: \texttt{Pendulum-v1} ($10000$ steps)}
    \label{fig:benchmark_sac}
  \end{minipage}
\vspace{-5.5mm}
\end{figure}
 \textbf{Horizontal Benchmark} \hspace{0.1cm} Figure \ref{fig:benchmark_ppo_results_medium} and \ref{fig:benchmark_sac} show the resulting mean training times from running the \gls*{ppo} and \gls*{sac} algorithm across ten runs on an Intel-based laptop (details in Table \ref{table:pendulum_sac_devices}). We find that RLtools outperforms existing libraries by a wide margin. Particularly in the case of \gls*{ppo} where RLtools only takes $\SI{0.54}{\second}$ on average ($\SI{2.59}{\second}$ in case of \gls*{sac}). 

\begin{table}
\vspace{1mm}
\begin{adjustbox}{center}
\resizebox{0.9\textwidth}{!}{
\def\arraystretch{1.3}
\begin{tabular}{c c c c}
\hline
 Platform & \rltools{}: Generic & DSP Library & \rltools{}: Optimized \\
\hline
\grayrow
Crazyflie & 743 us (1.3 kHz) & 478 us (2.1 kHz) & \textbf{293 us (3.4 kHz)}\\
Pixhawk 6C & 133 us (7.5 kHz) & 93 us (10.8 kHz) & \textbf{53 us (18.8 kHz)}\\
\grayrow
Teensy 4.1 & 64 us (15.5 kHz) & 45 us (22.3 kHz) & \textbf{41 us (24.3 kHz)}\\  
ESP32 (Xtensa) & 4282 us (234 Hz) & \textbf{279 us (3.6 kHz)} & 333 us (3 kHz)\\
\grayrow
ESP32-C3 (RISC-V) & 8716 us (115 Hz) & 6950 us (144 Hz) & \textbf{6645 us (150 Hz)}\\
\end{tabular}
}
\end{adjustbox}
\vspace{-4mm}
\caption{Inference times on different platforms}
\label{table:benchmark_inference}
\vspace{-5.5mm}
\end{table}

\begin{wrapfigure}{r}{0.50\textwidth}
    \vspace{-3mm}
  \centering
    \includegraphics[width=\linewidth]{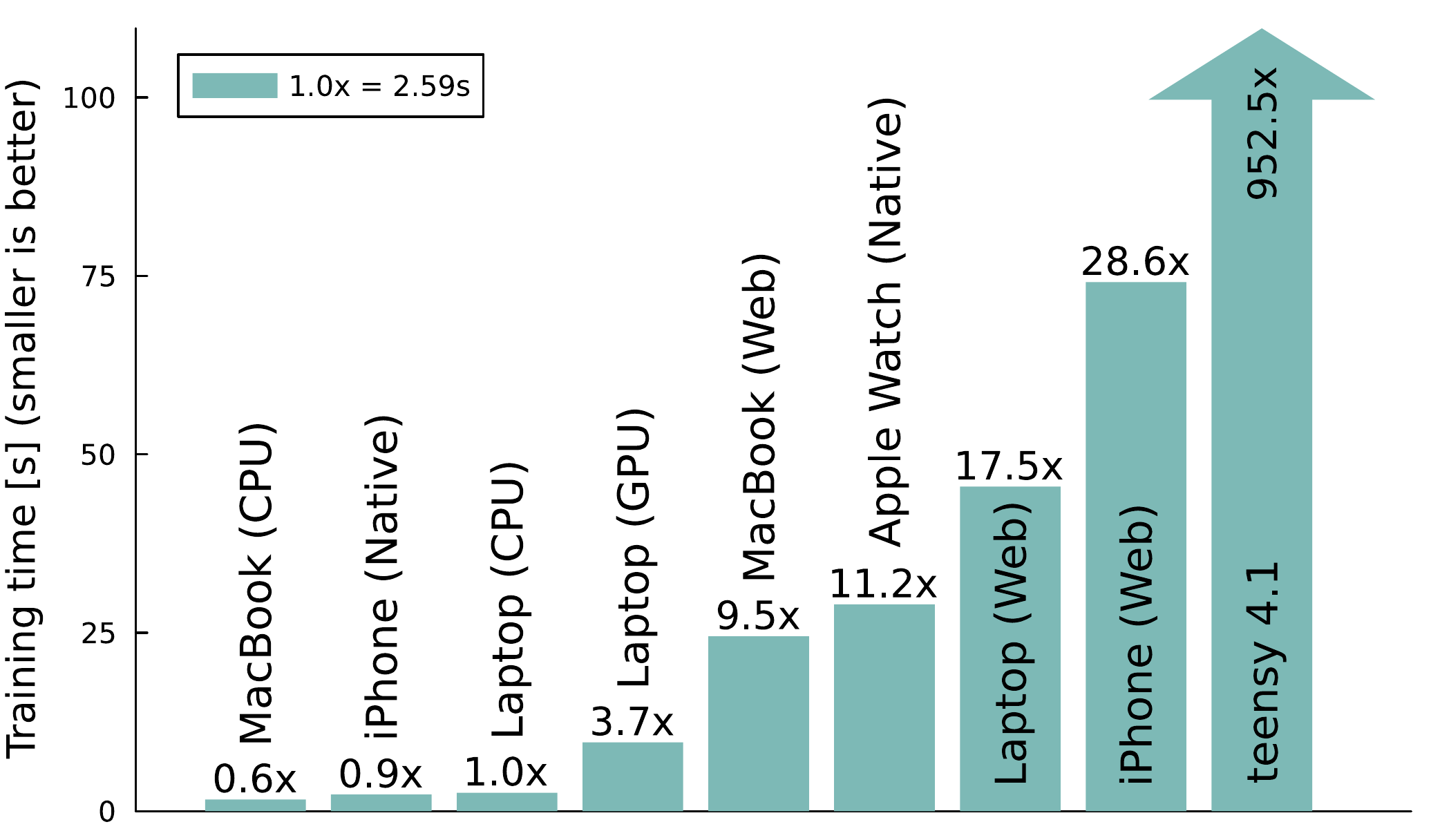}
    \centering
    \vspace{-10mm}
    \caption{\gls*{sac}: \texttt{Pendulum-v1} ($10000$ steps)}
    \label{fig:benchmark_sac_vertical}
    \vspace{-3mm}
\end{wrapfigure}

\noindent\textbf{Vertical Benchmark} \hspace{0.1cm} In Figure \ref{fig:benchmark_sac_vertical}, we also present training results using \rltools{} on a wide variety of devices which are generally not compatible with the other \gls*{rl} libraries and frameworks.  Importantly, we also demonstrate the first training of a deep \gls*{rl} agent for continuous control on a microcontroller in form of the Teensy 4.1.

\noindent\textbf{Inference on Microcontrollers}
Table \ref{table:benchmark_inference} shows the inference times on microcontrollers of different compute capabilities (e.g. Crazyflie is a \SI{27}{\gram} quadrotor with very limited resources, cf. Appendix \ref{appendix:embedded_platforms}). The generic implementation already yields usable inference times but dispatching to the manufacturers \gls*{dsp} library improves the performance. Finally, by optimizing the code further (e.g. through fusing the activation operators) we achieve a significant speedup even compared to the manufacturers \gls*{dsp} libraries.

\FloatBarrier
\section{Conclusion}
We believe \rltools{} fills a gap by allowing fast iteration in the problem space and subsequent real-time deployment of policies. Furthermore, \rltools{} facilitates the first-ever deep \gls*{rl} training on a microcontroller. 
We acknowledge the steeper learning curve of C++ (over e.g. Python) but from our experience, the faster iteration made possible by shorter training times 
 can outweigh the added time to get started. Currently \rltools{} is limited to dense observations but we plan to add vision capabilities in the future.
We believe that by relaxing the compute requirements and, by being fully open-source, \rltools{} democratizes the training of state-of-the-art \gls*{rl} methods and accelerates progress in \gls*{rl} for continuous control.

\FloatBarrier
\acks{
This work was supported by the Technology Innovation Institute, the NSF CAREER Award 2145277, and the DARPA YFA Grant D22AP00156-00. Giuseppe Loianno serves as a consultant for the Technology Innovation Institute. This arrangement has been reviewed and approved by New York University in accordance with its policy on objectivity in research.
}
\appendix
\FloatBarrier
\section{Analysis of the Deep RL Landscape}
\begin{table}[ht]
\def\arraystretch{1.3}
\centering
\begin{tabular}{c c c c c} 
 \hline
 Year & Name & Hidden Dim & \#Params & Non-Linearity \\
 \hline
\grayrow
 2015 & TRPO     \cite{schulman_trust_2015}            & [50, 50]        &   1.0x &  tanh\\
 2015 & GAE      \cite{schulman_high-dimensional_2015} & [100, 50, 25]   &   2.3x &  tanh\\
\grayrow
 2016 & DDPG     \cite{lillicrap_continuous_2016}      & [400, 300]      &  35.3x &  ReLU\\
 2017 & PPO      \cite{schulman_proximal_2017}         & [64, 64]        &   1.5x &  tanh\\
\grayrow
 2018 & TD3      \cite{fujimoto_addressing_2018}       & [400, 300]      &  35.3x &  ReLU\\
 2018 & SAC      \cite{haarnoja_soft_2018}             & [256, 256]      &  19.6x &  ReLU\\
\grayrow
 2019 & SACv2    \cite{haarnoja_soft_2019}             & [256, 256]      &  19.6x &  ReLU\\
 2020 & TQC      \cite{kuznetsov_controlling_2020}             & [512, 512, 512] & 147.0x &  ReLU\\
\grayrow
 2020 & D4PG\&TD3\cite{nicosia_learn_2020}     & [256, 256]      &  19.6x & ReLU\\
 2021 & PPO\&RMA  \cite{kumar_rma_2021}                & [128, 128, 128] &   9.8x & ReLU\\
 \hline
\end{tabular}
\caption{Selection of works that introduced impactful algorithms and the respective neural network dimensions used for their value function approximations. For the calculation of the number of parameters, an input size of $20$ and an output size of $1$ is assumed}
\label{table:nn_structures}
\end{table}
\label{appendix:analysis_drl}
In this section, we analyze the function approximator models used in the major deep \gls*{rl} for continuous control publications collected in Table \ref{table:deep_rl_framework_overview}. 
The most important observation is that over all the years the architecture (small, fully-connected neural networks) has not changed. 
This can be attributed to the fact that in continuous control the observations are usually dense states of the systems which do not contain any spatial or temporal regularities like images or time series that would suggest the usage of less general, more tailored network structures like \Glspl{cnn} or \Glspl{rnn}. This regularity, as stated in section \ref{sec:approach_architecture}, motivates our focus on optimizing and tightly integrating fully-connected neural networks as a first step. We also plan integrate recurrent and possibly convolutional layers in the future. 

\FloatBarrier
\section{Programming Paradigm}
\label{appendix:approach_programming_paradigm}
\definecolor{codegreen}{rgb}{0,0.6,0}
\definecolor{codegray}{rgb}{0.3,0.3,0.3}
\definecolor{codepurple}{rgb}{0.58,0,0.82}
\definecolor{backcolour}{rgb}{0.95,0.95,0.92}

\lstdefinestyle{bpt_style}{
    identifierstyle=\color{black},
    commentstyle=\color{codegray},
    keywordstyle=\color{primary_color_readable},
    numberstyle=\tiny\color{codegray},
    stringstyle=\color{black},
    basicstyle=\ttfamily\fontfamily{pcr}\selectfont\footnotesize,
    breakatwhitespace=false,         
    breaklines=true,                 
    captionpos=b,                    
    keepspaces=true,                 
    numbersep=5pt,                  
    showspaces=false,                
    showstringspaces=false,
    showtabs=false,                  
    tabsize=2,
}

\lstset{style=bpt_style}
\definecolor{listingbackgroundcolor}{HTML}{f5f5f5}
\begin{figure*}[t]
    \centering
    \begin{adjustwidth}{-0.05\textwidth}{-0.05\textwidth}
    \begin{mdframed}[roundcorner=10pt, linewidth=0, backgroundcolor=listingbackgroundcolor, userdefinedwidth=1.1\textwidth, align=center]
\begin{lstlisting}[language=C++]
// file: implementation_generic.h
template <typename DEVICE, auto M, auto N, auto K>
void multiply(DEVICE device, Matrix<M, K> a, Matrix<K, N> b, Matrix<M, N> result){
    // Generic code for matrix multiplication
    ...
}
// file: implementation_microcontroller.h
template <auto M, auto N, auto K>
void multiply(MICROCONTROLLER device, Matrix<M, K> a, Matrix<K, N> b, Matrix<M, N> result){
    // Optimized code for matrix multiplication on a particular microcontroller (e.g. based on DSP extensions)
    ...
}
// file: implementation_gpu.h
template <auto M, auto N, auto K>
void multiply(GPU device, Matrix<M, K> a, Matrix<K, N> b, Matrix<M, N> result){
    // Optimized GPU code for matrix multiplication (e.g. CUDA kernel launch)
    ...
}
template<typename DEVICE, typename OBJECT_A, typename OBJECT_B, typename OBJECT_C>
void algorithm(DEVICE device, OBJECT_A a, OBJECT_B b, OBJECT_C c){
    ...
    multiply(device, a, b, c);
    ...
}

// usage
GPU device;
Matrix<10, 10> a, b, result;
... // malloc and initialize randomly on the GPU device using tag dispatch as well
algorithm(device, a, b, result);
\end{lstlisting}
    \end{mdframed}
    \end{adjustwidth}
    \caption{Toy example for tag dispatch towards different implementations of elementary matrix operations}
    \label{fig:multiple_dispatch_code}
\end{figure*}

To enable maximum performance, we are avoiding C++ \gls*{vmt} lookups by not using an object-oriented paradigm but a rather functional paradigm heavily based on templating and method overloading resembling a static, compile-time defined interpretation of the multiple dispatch paradigm. Multiple dispatch has been popularized by the Julia programming language \cite{bezanson_julia_2012} and is based on advanced function overloading. 

Leveraging multiple dispatch, higher-level functions like    the forward or backward pass of a fully-connected neural network just specify the actions that should be taken on the different sub-components/layers and the actual implementation used is dependent on the type of the arguments. In this way, it is simple to share code between GPU and CPU implementations by just implementing the lower-level primitives for the respective device types and then signaling the implementations through the argument type (i.e. using the tag dispatch technique). A toy example for this is displayed in Figure \ref{fig:multiple_dispatch_code}. In this case, some \texttt{algorithm} is using a matrix multiplication operation on two objects. During the implementation of the \texttt{algorithm}, we do not need to care about the type of the operands and just let them be specified by wildcard template parameters. When this function is called by the user, the compiler infers the template parameters and dispatches the call to the appropriate implementation. If the user does not have a GPU available he simply does not include the \texttt{implementation\_gpu.h} and hence has no dependency on further dependencies that the GPU implementation would entail (e.g., the CUDA toolkit). In the case where there is no specialized implementation for a particular hardware, the compiler will fall back to the generic implementation which in this example could simply consist of a nested loop. The generic implementations are pure C++ and are guaranteed to have no dependencies. We can also see that the compiler will check the dimensions of the operands automatically at compile time such that the \texttt{algorithm} can not be called with incompatible shapes. To create more complex dispatch behaviors and operand type checking C++ features like \texttt{static\_assert} and \texttt{enable\_if} can be leveraged through the \gls*{sfinae} mechanism. In this way, we can maintain composability while still providing all the structure to the compiler at compile-time. 

In the case of Julia, this leads to unparalleled composability which manifests in a small number of people being able to implement and maintain e.g. a deep learning library (Flux \cite{innes_flux_2018}) that is competitive with PyTorch and TensorFlow which are backed by much more resources. 
In contrast to Julia, which reaches almost native performance while performing the multiple dispatch resolution at runtime, we make sure that all the function calls can be resolved at compile time. Additionally, Julia is not suited for our purposes because it does not fit to run on microcontrollers due to its runtime size and stochastic, non-realtime behavior due to the garbage collection-based memory management. Nevertheless, in our benchmark presented later in this manuscript, we found that Julia is one of the closest competitors when it comes to training performance.
Furthermore, we find it important to emphasize that we focus on building a library not a framework.\footnote{ \href{https://web.archive.org/web/20220614093124/https://www.brandons.me/blog/libraries-not-frameworks}{\textbf{Write Libraries, Not Frameworks} [link]}} The main feature of frameworks is that they restrict the freedoms of the user to make a small set of tasks easier to accomplish. In certain, repetitive problem settings this might be justified, but in many cases, the overhead coming with the steep learning curves and finding workarounds after bumping into the tight restrictions of frameworks is not worth it. The major conceptual difference is that frameworks provide a context from which they invoke the user's code while in the case of libraries, the user is entirely in control and invokes the components he needs. If not specifically made interoperable, the contexts provided by frameworks are usually incompatible while with libraries this is not generally the case. 

In our implementation, this for example concretely manifests in the way function approximators are used in the \gls*{rl} algorithms. By using templating, any function approximator type can be specified by the user at compile time. As long as he also provides the required \texttt{forward} and \texttt{backward} functions. 

As demonstrated in Figure \ref{fig:multiple_dispatch_code} we establish the convention of making a device-dependent context available in each function via tag dispatch to simplify the usage of different compute devices like accelerators or microcontrollers.

\FloatBarrier
\newpage
\section{Benchmark Details}
\label{appendix:benchmark_details}

\begin{table}[h]
\def\arraystretch{1.3}
\centering
\vspace{2mm}
\begin{tabular}{c c} 
 \hline
 Parameter & Value \\
 \hline
\grayrow
 Actor structure & $[64, 64]$ \\
 Critic structure & $[64, 64]$ \\
\grayrow
 Activation function & \gls*{relu} \\
 Batch size & $256$\\
\grayrow
 Number of environments & $4$ \\
 Steps per environment & $1024$ \\
\grayrow
 Number of epochs & 2 \\
 Total number of (environment) steps & 300000 \\
\grayrow
 Discount factor $\gamma$ & $0.9$\\
 \gls*{gae} $\lambda$ & $0.95$\\
\grayrow
 $\epsilon$ clip & $0.2$\\
 Entropy coefficient $\beta$  & $0$\\
\grayrow
 Advantage normalization & \texttt{true}\\
 Adam $\alpha$ & $1 \times 10^{-3}$\\
\grayrow
 Adam $\beta_1$ & $0.9$\\
 Adam $\beta_2$ & $0.999$\\
\grayrow
 Adam $\epsilon$ & $1 \times 10^{-7}$\\
 \hline
\end{tabular}
\caption{\texttt{Pendulum-v1} \gls*{ppo} parameters (Figure \ref{fig:benchmark_ppo_results_medium})}
\label{table:pendulum_ppo_parameters}
\end{table}

\begin{table}[h]
\def\arraystretch{1.3}
\centering
\vspace{2mm}
\begin{tabular}{c c} 
 \hline
 Parameter & Value \\
 \hline
\grayrow
 Actor structure & $[64, 64]$ \\
 Critic structure & $[64, 64]$ \\
\grayrow
 Activation function & \gls*{relu} \\
 Batch size & $100$\\
\grayrow
 Total number of (environment) steps & $10000$ \\
 Replay buffer size & $10000$ \\
\grayrow
 Discount factor $\gamma$ & $0.99$\\
 Entropy bonus coefficient (learned, initial value) $\alpha$ & $0.5$\\
\grayrow
 Polyak $\beta$ & $0.99$\\
 Adam $\alpha$ & $1 \times 10^{-3}$\\
\grayrow
 Adam $\beta_1$ & $0.9$\\
 Adam $\beta_2$ & $0.999$\\
\grayrow
 Adam $\epsilon$ & $1 \times 10^{-7}$\\
 \hline
\end{tabular}
\caption{\texttt{Pendulum-v1} \gls*{sac} parameters (Figure \ref{fig:benchmark_sac})}
\label{table:pendulum_sac_parameters}
\end{table}

\begin{table}[h]
\def\arraystretch{1.3}
\centering
\vspace{2mm}
\begin{tabular}{c c} 
 \hline
 Parameter & Value \\
 \hline
\grayrow
 Input dimensionality & $13$ \\
 Policy structure & $[64, 64]$ \\
\grayrow
 Output dimensionality & $4$ \\
 Activation function & \gls*{relu}\\
 \hline
\end{tabular}
\caption{On-device inference parameters (Table \ref{table:benchmark_inference})}
\label{table:benchmark_inference_parameters}
\end{table}

\begin{table}[ht]
\def\arraystretch{1.3}
\centering
\vspace{2mm}
\resizebox{\textwidth}{!}{
\begin{tabular}{c c c c c} 
 \hline
 Label & Details \\
 \hline
\grayrow
 \rltools{} / Laptop (CPU) / Laptop (Web) / Baseline & Intel i9-10885H \\
 Laptop (GPU) & Intel i9-10885H + Nvidia T2000 \\
\grayrow
 MacBook (CPU) / MacBook (Web) & MacBook Pro (M3 Pro) \\
 iPhone (Native) / iPhone (Web) & iPhone 14 \\
\grayrow
 Apple Watch (Native) & Apple Watch Series 4\\
 \hline
\end{tabular}
}
\caption{\texttt{Pendulum-v1} \gls*{sac} devices (Figure \ref{fig:benchmark_sac_vertical})}
\label{table:pendulum_sac_devices}
\end{table}
\FloatBarrier
\section{Deep Reinforcement Learning Frameworks and Libraries}
\label{appendix:deep_rl_frameworks_and_libraries}
\begin{table}[h]
\def\arraystretch{1.3}
\centering
\begin{tabular}{c  c  c}
\hline
Name \raisebox{1pt}{$\downarrow$} & Platform & Stars / Citations \\
\hline
\grayrow
Acme \citep{hoffman_acme_2022}                         & JAX           & 3316  / 219\\
CleanRL \citep{huang_cleanrl_2021}                     & PyTorch       & 4030  / 86\\
\grayrow
MushroomRL \citep{deramo_mushroomrl_2020}              & TF/PyTorch & 749  / 61\\
PFRL \citep{fujita_chainerrl_2021}                     & PyTorch       & 1125  / 122\\
\grayrow
ReinforcementLearning.jl (JuliaRL) \citep{Tian2020Reinforcement} & Flux.jl (Julia)           & 543  / n/a\\
RLlib + \texttt{ray} \citep{liang_rllib_2018}          & PyTorch       & 29798 / 828\\
\grayrow
Stable Baselines3 (SB3) \citep{ran_stable-baselines3_2021}   & PyTorch       &  7396 / 1149\\
Stable Baselines JAX (SBX) \citep{ran_stable-baselines3_2021}   & JAX       &  223 / n/a\\
\grayrow
Tianshou \citep{weng_tianshou_2022}                    & PyTorch       & 7139  / 133\\
TorchRL \citep{weng_tianshou_2022}                     & PyTorch       & 1691   / 5\\
\hline
\end{tabular}
\caption{Overview over different \Gls{rl} libraries/frameworks, the deep learning platform they build upon, and their popularity in terms of Github stars and publication citations (data as of 2024-02-07)}
\label{table:deep_rl_framework_overview}
\end{table}
\FloatBarrier
\section{Embedded Platforms}
\label{appendix:embedded_platforms}
\begin{enumerate}
\item \textbf{Crazyflie}: A small, open-source quadrotor which only weighs $27$ g including the battery. The Crazyflie's main processor is a STM32F405 microcontroller using the ARM Cortex-M4 architecture, featuring $192$ KB of \gls*{ram} and running at $168$ MHz. 

\item \textbf{Pixhawk 6C}: We use a Pixracer Pro, a \gls*{fcu} that belongs to the family of Pixhawk \Glspl{fcu} and implements the Pixhawk 6C standard. Hence, the PixRacer Pro supports the common PX4 firmware \cite{7140074} and can be used in many different vehicle types (aerial, ground, marine) but is predominantly used in multirotor vehicles of varying sizes. The main processor used in the Pixhawk 6C standard is a STM32H743 using the ARM Cortex-M7 architecture. The PixRacer Pro runs at $460$ MHz and comes with $1024$ KB of \gls*{ram}. 

\item \textbf{Teensy 4.1}: A general-purpose embedded device powered by an i.MX RT1060 ARM Cortex-M7 microcontroller with $1024$ KB on-chip and $16$ MB off-chip \gls*{ram} that is running at $600$ MHz.

\item \textbf{ESP32}: One of the most common microcontrollers for \gls*{iot} and edge devices due to its built-in Wi-Fi and Bluetooth. Close to \num{1} billion devices built around this chip and its predecessor have been sold worldwide. Hence it is widely available and relatively cheap (around $\$$5 for a development kit). For our purposes, the ESP32 is interesting because it deviates from the previous platforms in that its processor is based on the Xtensa LX7 architecture. In addition to the original version of the ESP32 based on the Xtensa architecture, we also evaluate the ESP32-C3 version based on the RISC-V architecture.
\end{enumerate}
\FloatBarrier
\section{Ablation Study}
\label{appendix:ablation_study}
\FloatBarrier
\begin{wrapfigure}{r}{0.50\textwidth}
% \begin{figure}[h]
\includegraphics[width=\linewidth]{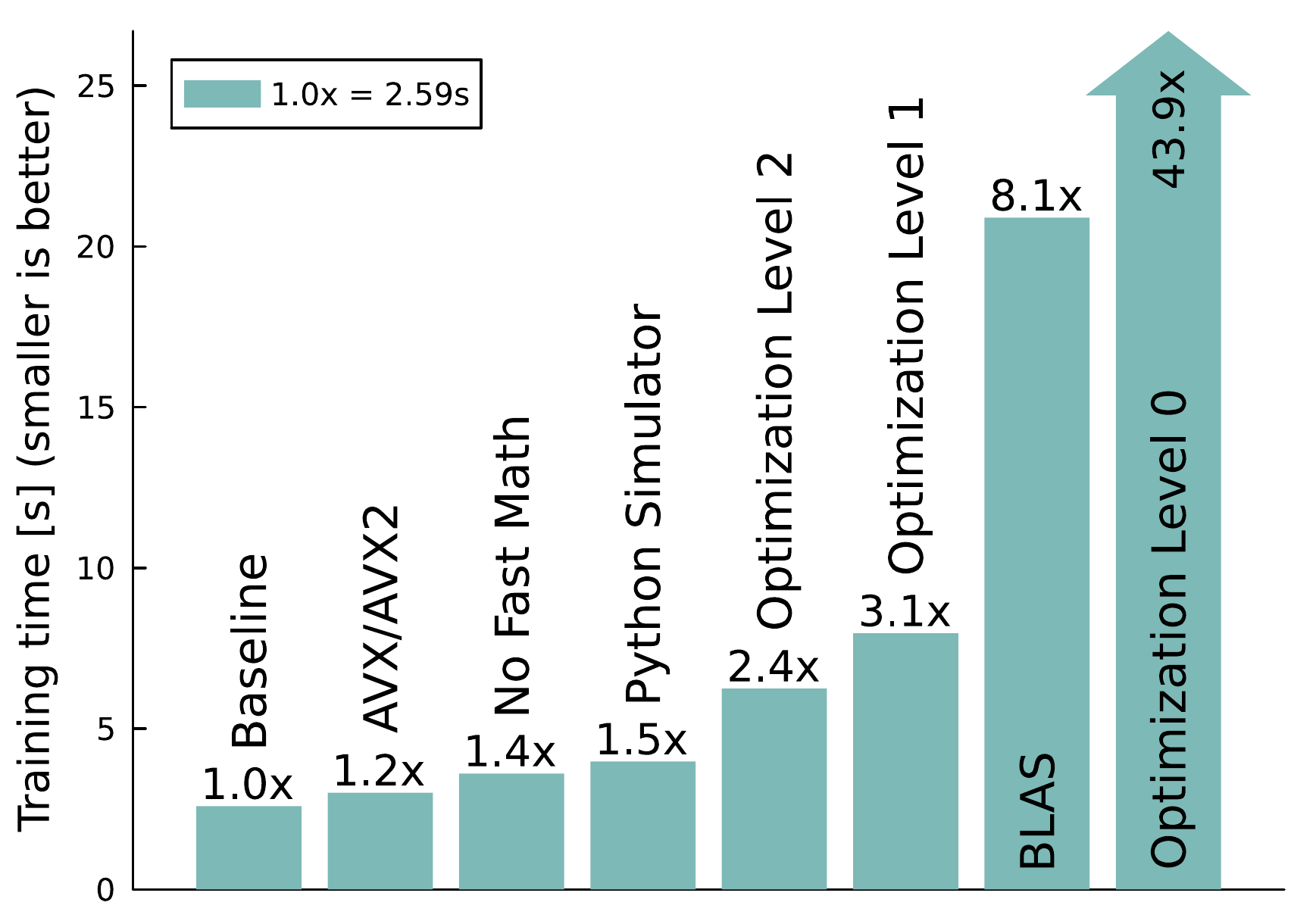}
\centering
\caption{Ablation study. The ``Baseline'' contains all optimizations.}
\label{fig:ablation_study}
% \end{figure}
\end{wrapfigure}
We conduct an ablation study to investigate the contribution of different components and optimizations to the fast wall-clock training time achieved by \rltools{}. Figure \ref{fig:ablation_study} shows the resulting training times after removing different components and optimizations from the setup. The ``Baseline'' is exactly the same setup used in the \texttt{Pendulum-v1} (\gls*{sac}) training in the other experiments in Section \ref{sec:results}. We simulate the slowness of the Python environment by slowing down the C++ implementation by the average time required for a step in the Python implementation. We can observe that the C++ implementation of the \texttt{Pendulum-v1} dynamics has a measurable, but not dominating impact on the training time. Additionally, we ablate the different optimization levels \texttt{-O0}, \texttt{-O1}, \texttt{-O2} and \texttt{-O3} (used in the Baseline) of the C++ compiler. We can observe that the compiler optimizations have a sizable impact on the training time. When removing all optimizations (\texttt{-O0}) \rltools{} is roughly between ACME and CleanRL (cf. Figure \ref{fig:benchmark_sac}). Furthermore, the ``No Fast Math'' configuration tests removing the \texttt{-ffast-math} from the compiler options, and the ``BLAS'' \gls*{blas} option removes the Intel oneMKL matrix multiplication kernels. In the case of ``AVX/AVX2'' we disable the \gls*{avx} that are used for \gls*{simd} operations. We notice that due to the design of \rltools{} (refer to Appendix \ref{appendix:approach_programming_paradigm}) which allows the sizes of all loops and data structures to be known at compile-time the compiler is able to better reason about the code and hence make heavy use of vectorized/\gls*{simd} operations. We observe that $2276 + 1430 = 3706$ (\gls*{avx} + \gls*{sse}, an older set of vectorized instructions) out of $11243$ machine-code instructions in total refer to registers of the vector extensions. Unfortunately (for the sake of measurement), when turning off \gls*{avx}, the compiler replaces the instructions with \gls*{sse} instructions ($5406$ out of $11243$ in this case) which we could not turn off because of some dependency in \texttt{libstdc++}. Still, the number of \gls*{sse} instructions demonstrates the compiler-friendliness that \rltools{}' architecture entails. 

\FloatBarrier
\section{Convergence Study}
\FloatBarrier

To make sure the implementations of the supported \gls*{rl} algorithms (PPO, TD3, and SAC) are correct, we conduct a convergence study where we compare the learning curves across different environments with learning curves of other implementations. We make sure that per environment the same hyperparameters are used across all implementations and run each setup for multiple seeds (\num{100} for \texttt{Pendulum-v1} and \num{30} for \texttt{Hopper-v1}). For each of the seeds at every evaluation step, we perform \num{100} episodes with random initial states.

By comparing different sets of \num{100} seeds each, we found that, even for a large number of seeds, outliers have a significant impact on the mean final return. Hence, as also recommended by \citet{agarwal2021rlprecipe}, we report the \gls*{iqm} which discards the lowest and highest quantile to remove the impact of outliers on the statistics. We still aim at capturing as much of the final return distribution by only discarding the lower upper \SI{5}{\percent} for the calculation of the \gls*{iqm} $\mu$. We use the same inter-quantile set for the calculation of the standard deviation $\sigma$. To make sure that the environments are identical in this convergence study, instead of re-implementing the environments in C++ using the \rltools{} interface, we built a Python wrapper for \rltools{} such that we can use the original environments from the Gymnasium \citep{Towers_Gymnasium} suite. The Python wrapper makes \rltools{} easier to use but sacrifices in terms of performance if the environment/simulator is implemented in Python (as shown in Appendix \ref{appendix:ablation_study}).

\begin{figure}[h]
\includegraphics[width=0.80\textwidth]{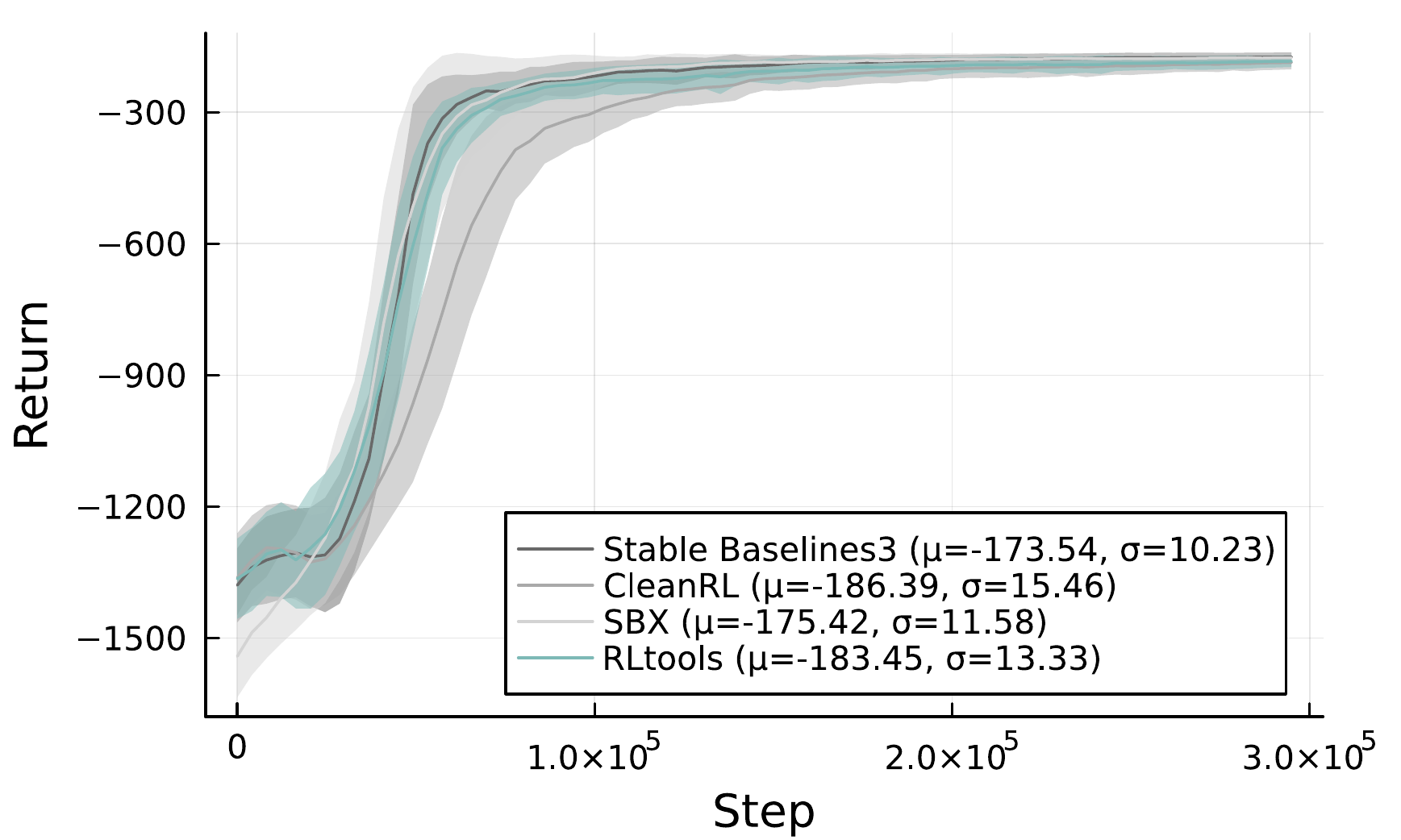}
\centering
\caption{PPO \texttt{Pendulum-v1}}
\label{fig:convergence_study_ppo_pendulum}
\end{figure}
\begin{figure}
\includegraphics[width=0.80\textwidth]{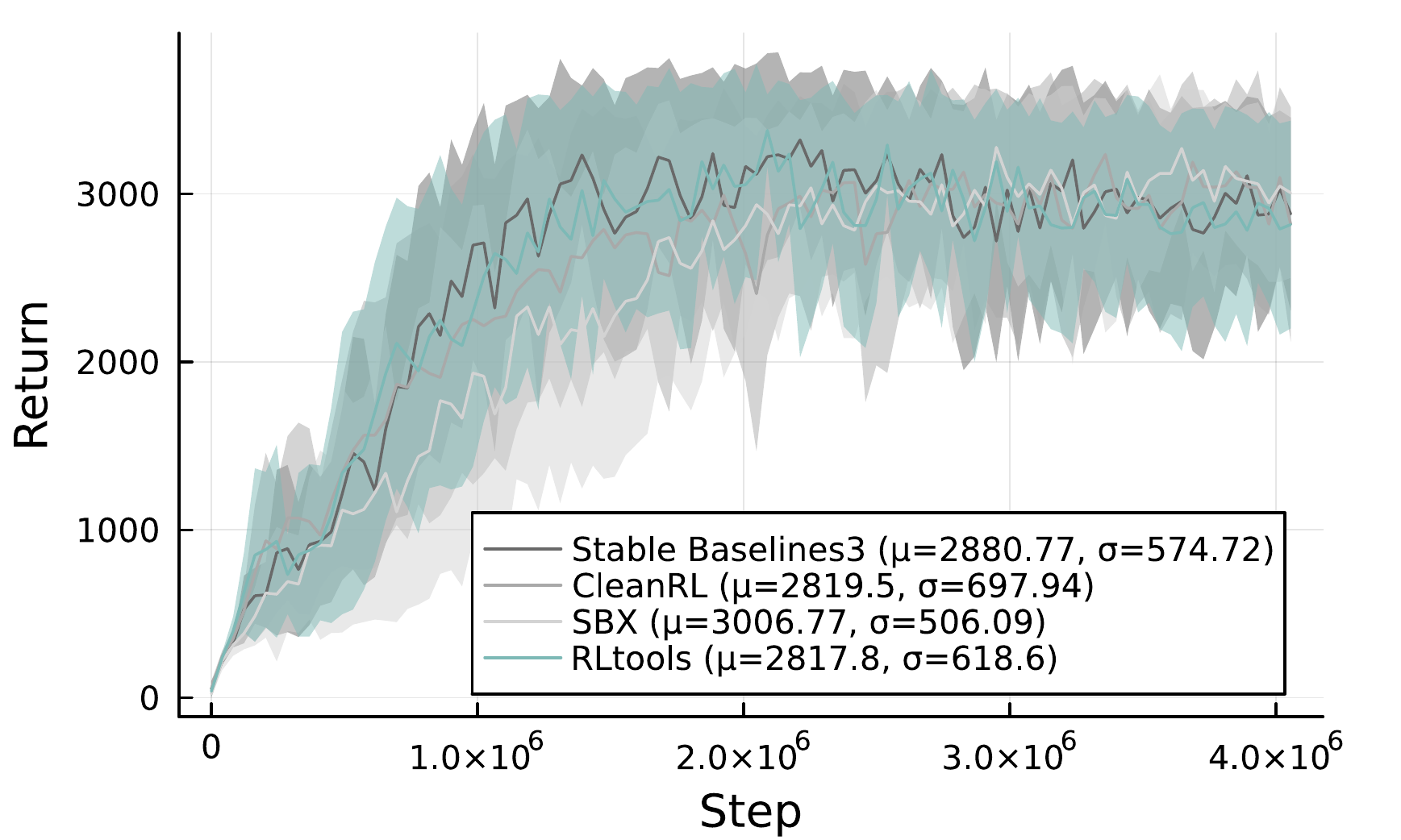}
\centering
\caption{PPO \texttt{Hopper-v4}}
\label{fig:convergence_study_ppo_hopper}
\end{figure}
\begin{figure}
\includegraphics[width=0.80\textwidth]{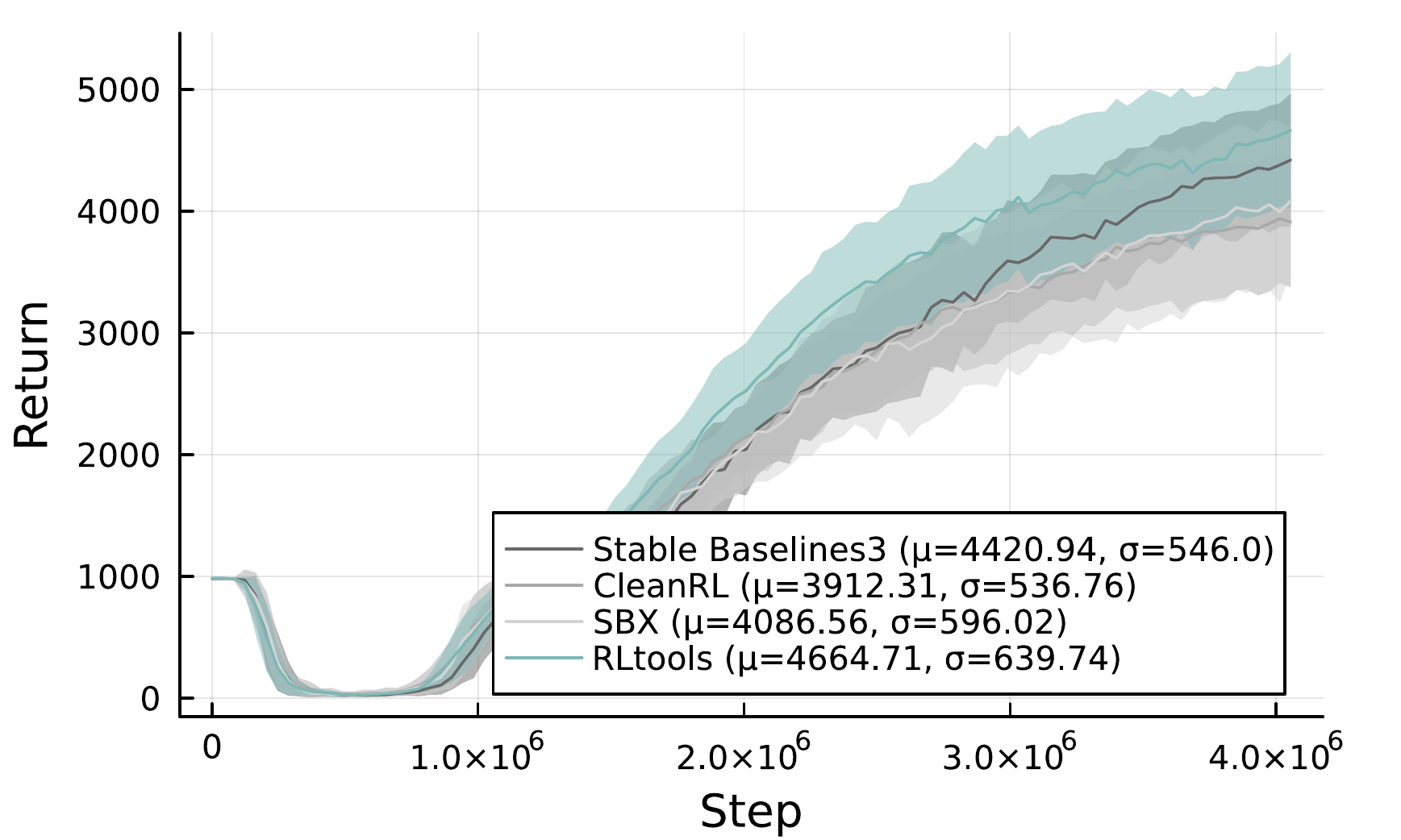}
\centering
\caption{PPO \texttt{Ant-v4}}
\label{fig:convergence_study_ppo_ant}
\end{figure}

\begin{figure}
\includegraphics[width=0.80\textwidth]{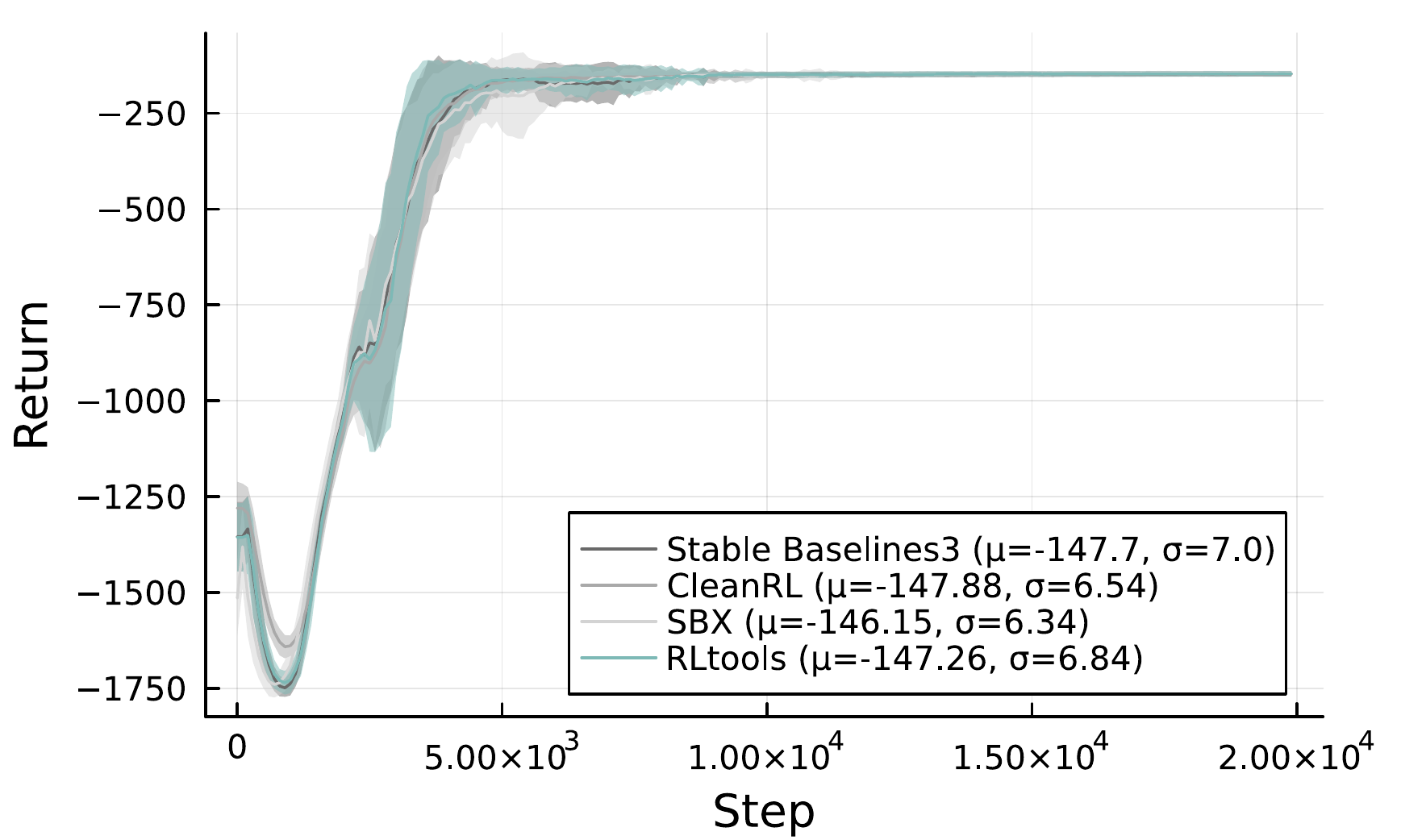}
\centering
\caption{SAC \texttt{Pendulum-v1}}
\label{fig:convergence_study_sac_pendulum}
\end{figure}
\begin{figure}
\includegraphics[width=0.80\textwidth]{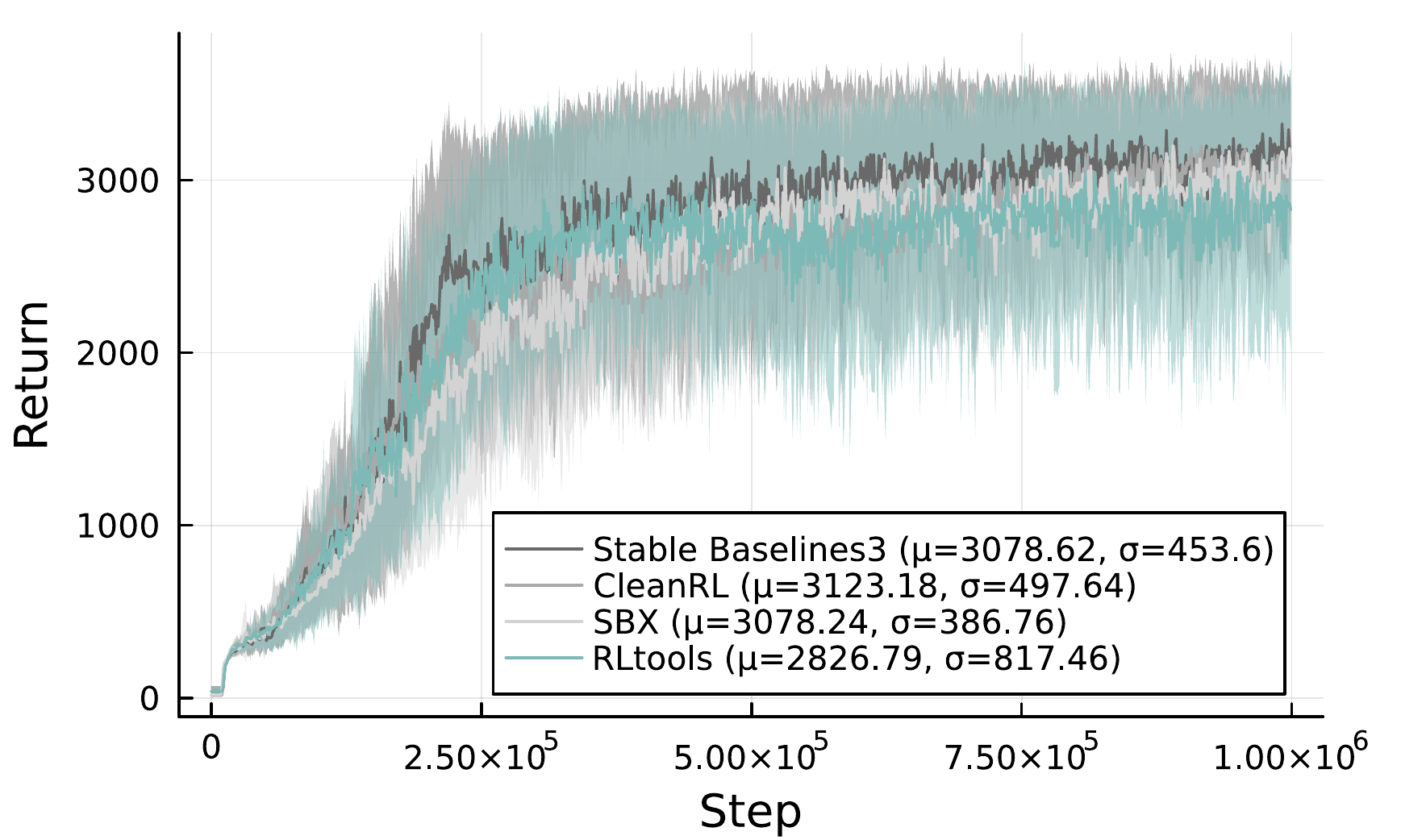}
\centering
\caption{SAC \texttt{Hopper-v4}}
\label{fig:convergence_study_sac_hopper}
\end{figure}
\begin{figure}
\includegraphics[width=0.80\textwidth]{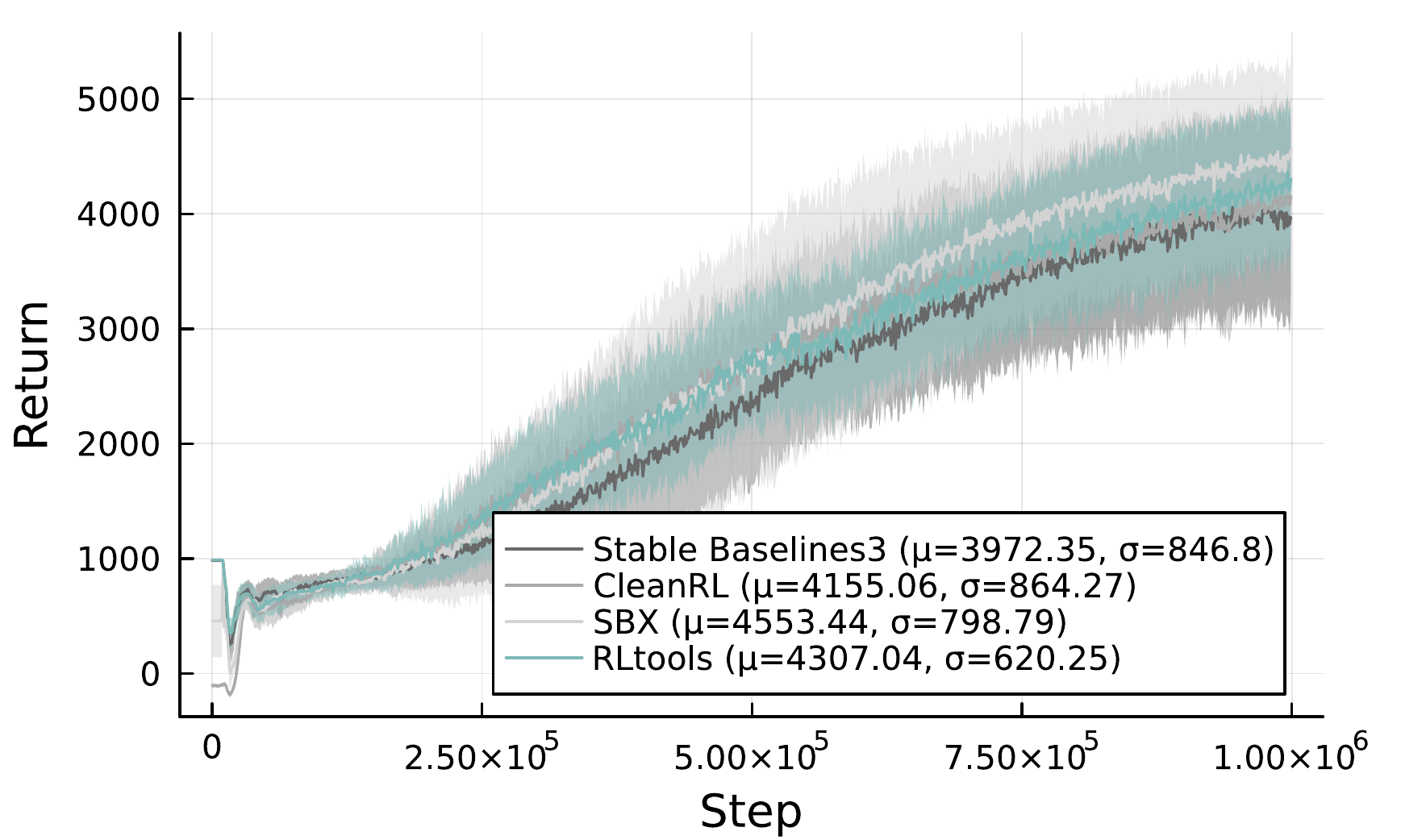}
\centering
\caption{SAC \texttt{Ant-v4}}
\label{fig:convergence_study_sac_ant}
\end{figure}

\begin{figure}
\includegraphics[width=0.80\textwidth]{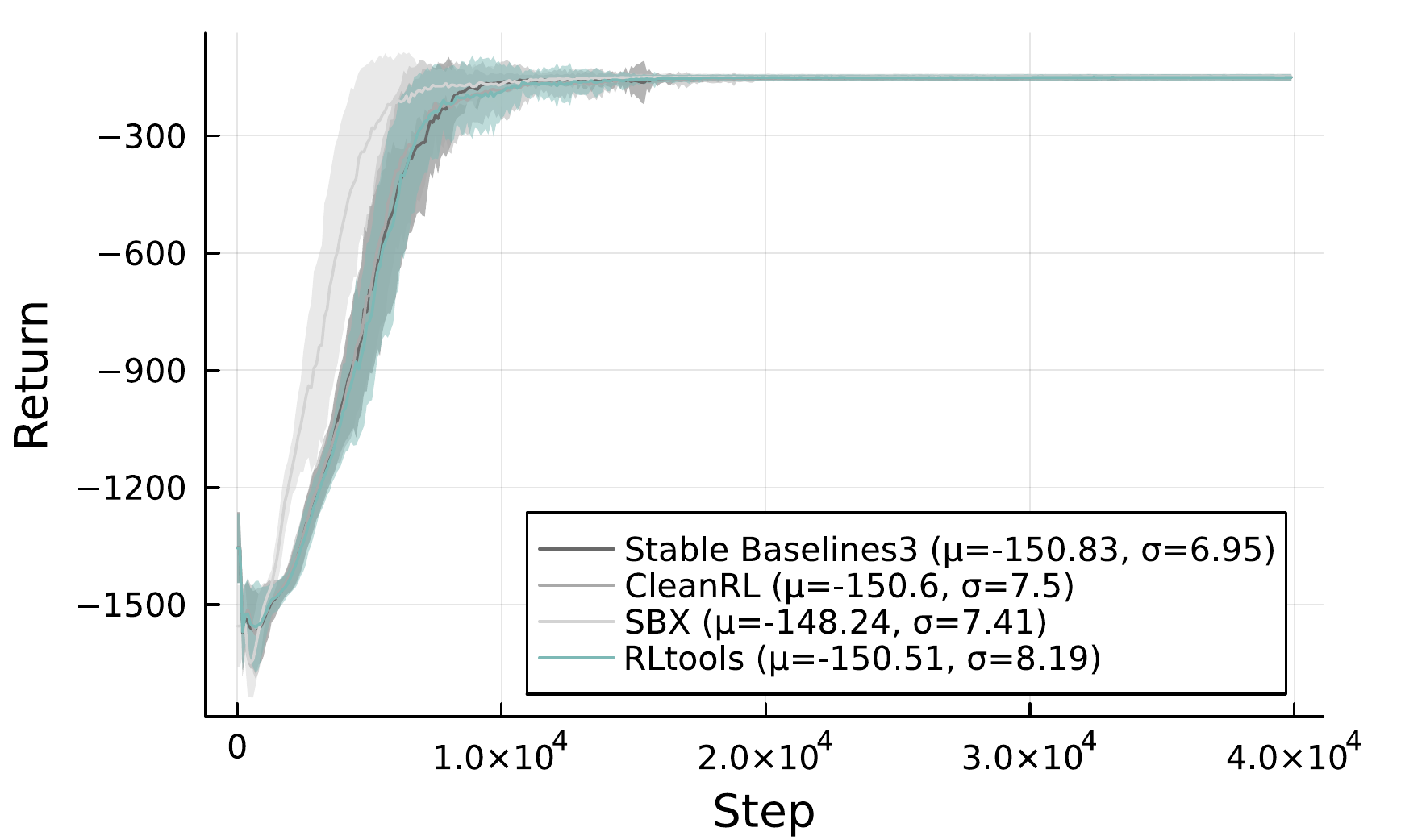}
\centering
\caption{TD3 \texttt{Pendulum-v1}}
\label{fig:convergence_study_td3_pendulum}
\end{figure}
\begin{figure}
\includegraphics[width=0.80\textwidth]{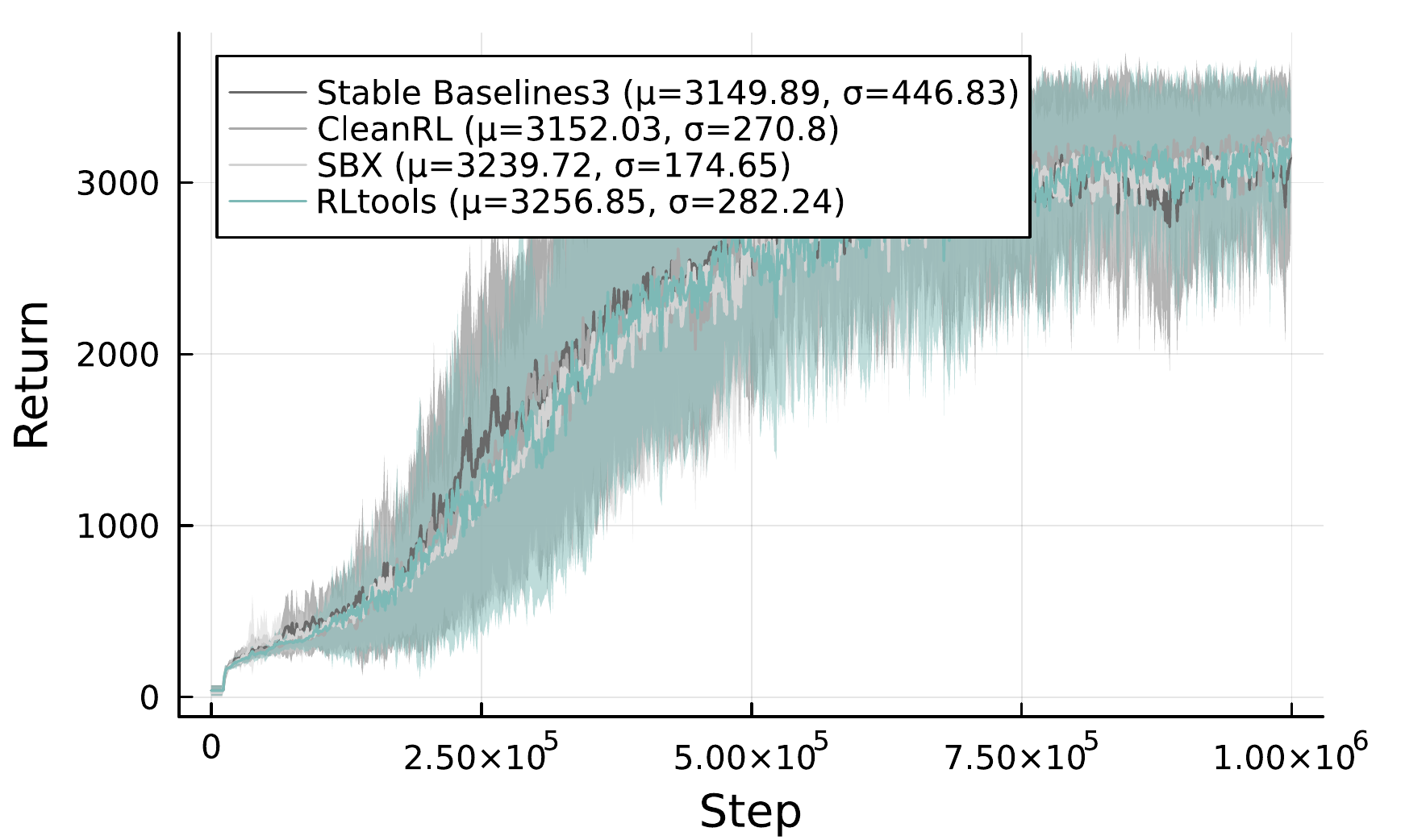}
\centering
\caption{TD3 \texttt{Hopper-v4}}
\label{fig:convergence_study_td3_hopper}
\end{figure}
\begin{figure}
\includegraphics[width=0.80\textwidth]{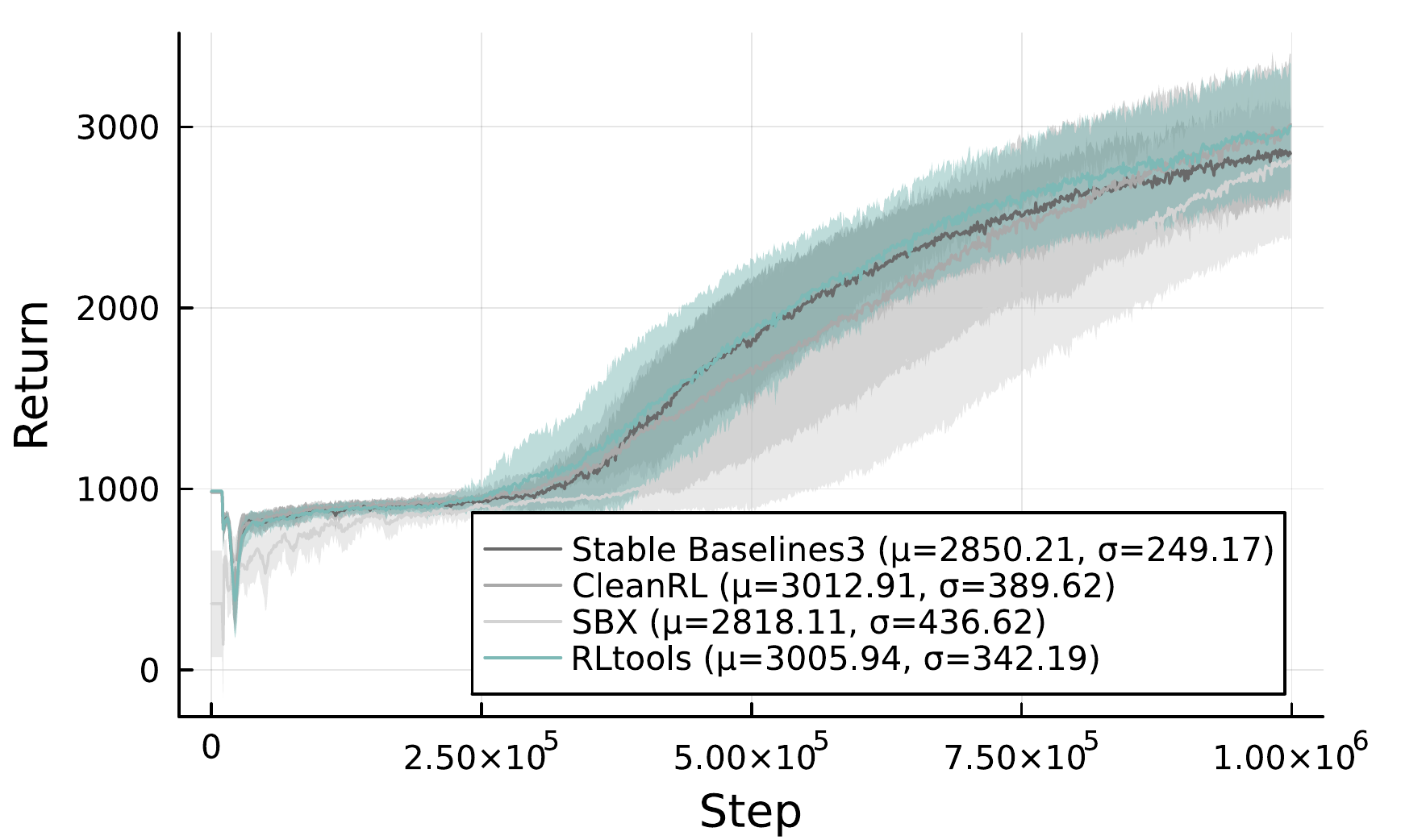}
\centering
\caption{TD3 \texttt{Ant-v4}}
\label{fig:convergence_study_td3_ant}
\end{figure}
\FloatBarrier
\FloatBarrier
\bibliography{additional_references}

\begin{thebibliography}{27}
\providecommand{\natexlab}[1]{#1}
\providecommand{\url}[1]{\texttt{#1}}
\expandafter\ifx\csname urlstyle\endcsname\relax
  \providecommand{\doi}[1]{doi: #1}\else
  \providecommand{\doi}{doi: \begingroup \urlstyle{rm}\Url}\fi

\bibitem[Agarwal et~al.(2021)Agarwal, Schwarzer, Castro, Courville, and Bellemare]{agarwal2021rlprecipe}
Rishabh Agarwal, Max Schwarzer, Pablo~Samuel Castro, Aaron~C Courville, and Marc Bellemare.
\newblock Deep reinforcement learning at the edge of the statistical precipice.
\newblock In M.~Ranzato, A.~Beygelzimer, Y.~Dauphin, P.S. Liang, and J.~Wortman Vaughan, editors, \emph{Advances in Neural Information Processing Systems}, volume~34, pages 29304--29320. Curran Associates, Inc., 2021.
\newblock URL \url{https://proceedings.neurips.cc/paper_files/paper/2021/file/f514cec81cb148559cf475e7426eed5e-Paper.pdf}.

\bibitem[Bach et~al.(2020)Bach, Melnik, Schilling, Korthals, and Ritter]{nicosia_learn_2020}
Nicolas Bach, Andrew Melnik, Malte Schilling, Timo Korthals, and Helge Ritter.
\newblock Learn to {Move} {Through} a {Combination} of {Policy} {Gradient} {Algorithms}: {DDPG}, {D4PG}, and {TD3}.
\newblock In Giuseppe Nicosia, Varun Ojha, Emanuele La~Malfa, Giorgio Jansen, Vincenzo Sciacca, Panos Pardalos, Giovanni Giuffrida, and Renato Umeton, editors, \emph{Machine {Learning}, {Optimization}, and {Data} {Science}}, volume 12566, pages 631--644. Springer International Publishing, Cham, 2020.
\newblock ISBN 978-3-030-64579-3 978-3-030-64580-9.
\newblock \doi{10.1007/978-3-030-64580-9_52}.
\newblock URL \url{http://link.springer.com/10.1007/978-3-030-64580-9_52}.
\newblock Series Title: Lecture Notes in Computer Science.

\bibitem[Bezanson et~al.(2012)Bezanson, Karpinski, Shah, and Edelman]{bezanson_julia_2012}
Jeff Bezanson, Stefan Karpinski, Viral~B. Shah, and Alan Edelman.
\newblock Julia: {A} {Fast} {Dynamic} {Language} for {Technical} {Computing}, September 2012.
\newblock URL \url{http://arxiv.org/abs/1209.5145}.
\newblock arXiv:1209.5145 [cs].

\bibitem[Bou et~al.(2024)Bou, Bettini, Dittert, Kumar, Sodhani, Yang, Fabritiis, and Moens]{bou2023torchrl}
Albert Bou, Matteo Bettini, Sebastian Dittert, Vikash Kumar, Shagun Sodhani, Xiaomeng Yang, Gianni~De Fabritiis, and Vincent Moens.
\newblock Torch{RL}: A data-driven decision-making library for pytorch.
\newblock In \emph{The Twelfth International Conference on Learning Representations}, 2024.
\newblock URL \url{https://openreview.net/forum?id=QxItoEAVMb}.

\bibitem[D'Eramo et~al.(2021)D'Eramo, Tateo, Bonarini, Restelli, and Peters]{deramo_mushroomrl_2020}
Carlo D'Eramo, Davide Tateo, Andrea Bonarini, Marcello Restelli, and Jan Peters.
\newblock Mushroomrl: Simplifying reinforcement learning research.
\newblock \emph{Journal of Machine Learning Research}, 22\penalty0 (131):\penalty0 1--5, 2021.
\newblock URL \url{http://jmlr.org/papers/v22/18-056.html}.

\bibitem[Eschmann(2021)]{eschmann2021reward}
Jonas Eschmann.
\newblock Reward function design in reinforcement learning.
\newblock \emph{Reinforcement Learning Algorithms: Analysis and Applications}, pages 25--33, 2021.

\bibitem[Fujimoto et~al.(2018)Fujimoto, van Hoof, and Meger]{fujimoto_addressing_2018}
Scott Fujimoto, Herke van Hoof, and David Meger.
\newblock Addressing function approximation error in actor-critic methods.
\newblock In Jennifer Dy and Andreas Krause, editors, \emph{Proceedings of the 35th International Conference on Machine Learning}, volume~80 of \emph{Proceedings of Machine Learning Research}, pages 1587--1596. PMLR, 10--15 Jul 2018.
\newblock URL \url{https://proceedings.mlr.press/v80/fujimoto18a.html}.

\bibitem[Fujita et~al.(2021)Fujita, Nagarajan, Kataoka, and Ishikawa]{fujita_chainerrl_2021}
Yasuhiro Fujita, Prabhat Nagarajan, Toshiki Kataoka, and Takahiro Ishikawa.
\newblock Chainerrl: A deep reinforcement learning library.
\newblock \emph{Journal of Machine Learning Research}, 22\penalty0 (77):\penalty0 1--14, 2021.
\newblock URL \url{http://jmlr.org/papers/v22/20-376.html}.

\bibitem[Haarnoja et~al.(2018)Haarnoja, Zhou, Abbeel, and Levine]{haarnoja_soft_2018}
Tuomas Haarnoja, Aurick Zhou, Pieter Abbeel, and Sergey Levine.
\newblock Soft actor-critic: Off-policy maximum entropy deep reinforcement learning with a stochastic actor.
\newblock In Jennifer Dy and Andreas Krause, editors, \emph{Proceedings of the 35th International Conference on Machine Learning}, volume~80 of \emph{Proceedings of Machine Learning Research}, pages 1861--1870. PMLR, 10--15 Jul 2018.
\newblock URL \url{https://proceedings.mlr.press/v80/haarnoja18b.html}.

\bibitem[Haarnoja et~al.(2019)Haarnoja, Zhou, Hartikainen, Tucker, Ha, Tan, Kumar, Zhu, Gupta, Abbeel, and Levine]{haarnoja_soft_2019}
Tuomas Haarnoja, Aurick Zhou, Kristian Hartikainen, George Tucker, Sehoon Ha, Jie Tan, Vikash Kumar, Henry Zhu, Abhishek Gupta, Pieter Abbeel, and Sergey Levine.
\newblock Soft {Actor}-{Critic} {Algorithms} and {Applications}, January 2019.
\newblock URL \url{http://arxiv.org/abs/1812.05905}.
\newblock arXiv:1812.05905 [cs, stat].

\bibitem[Hoffman et~al.(2020)Hoffman, Shahriari, Aslanides, Barth-Maron, Momchev, Sinopalnikov, Sta\'nczyk, Ramos, Raichuk, Vincent, Hussenot, Dadashi, Dulac-Arnold, Orsini, Jacq, Ferret, Vieillard, Ghasemipour, Girgin, Pietquin, Behbahani, Norman, Abdolmaleki, Cassirer, Yang, Baumli, Henderson, Friesen, Haroun, Novikov, Colmenarejo, Cabi, Gulcehre, Paine, Srinivasan, Cowie, Wang, Piot, and de~Freitas]{hoffman_acme_2022}
Matthew~W. Hoffman, Bobak Shahriari, John Aslanides, Gabriel Barth-Maron, Nikola Momchev, Danila Sinopalnikov, Piotr Sta\'nczyk, Sabela Ramos, Anton Raichuk, Damien Vincent, L\'eonard Hussenot, Robert Dadashi, Gabriel Dulac-Arnold, Manu Orsini, Alexis Jacq, Johan Ferret, Nino Vieillard, Seyed Kamyar~Seyed Ghasemipour, Sertan Girgin, Olivier Pietquin, Feryal Behbahani, Tamara Norman, Abbas Abdolmaleki, Albin Cassirer, Fan Yang, Kate Baumli, Sarah Henderson, Abe Friesen, Ruba Haroun, Alex Novikov, Sergio~G\'omez Colmenarejo, Serkan Cabi, Caglar Gulcehre, Tom~Le Paine, Srivatsan Srinivasan, Andrew Cowie, Ziyu Wang, Bilal Piot, and Nando de~Freitas.
\newblock Acme: A research framework for distributed reinforcement learning.
\newblock \emph{arXiv preprint arXiv:2006.00979}, 2020.
\newblock URL \url{https://arxiv.org/abs/2006.00979}.

\bibitem[Huang et~al.(2022)Huang, Dossa, Ye, Braga, Chakraborty, Mehta, and Araujo]{huang_cleanrl_2021}
Shengyi Huang, Rousslan Fernand~Julien Dossa, Chang Ye, Jeff Braga, Dipam Chakraborty, Kinal Mehta, and Joao~G.M. Araujo.
\newblock Cleanrl: High-quality single-file implementations of deep reinforcement learning algorithms.
\newblock \emph{Journal of Machine Learning Research}, 23\penalty0 (274):\penalty0 1--18, 2022.
\newblock URL \url{http://jmlr.org/papers/v23/21-1342.html}.

\bibitem[Innes(2018)]{innes_flux_2018}
Mike Innes.
\newblock Flux: {Elegant} machine learning with {Julia}.
\newblock \emph{Journal of Open Source Software}, 3\penalty0 (25):\penalty0 602, May 2018.
\newblock ISSN 2475-9066.
\newblock \doi{10.21105/joss.00602}.
\newblock URL \url{http://joss.theoj.org/papers/10.21105/joss.00602}.

\bibitem[Kumar et~al.(2021)Kumar, Fu, Pathak, and Malik]{kumar_rma_2021}
Ashish Kumar, Zipeng Fu, Deepak Pathak, and Jitendra Malik.
\newblock {RMA: Rapid Motor Adaptation for Legged Robots}.
\newblock In \emph{Proceedings of Robotics: Science and Systems}, Virtual, July 2021.
\newblock \doi{10.15607/RSS.2021.XVII.011}.

\bibitem[Kuznetsov et~al.(2020)Kuznetsov, Shvechikov, Grishin, and Vetrov]{kuznetsov_controlling_2020}
Arsenii Kuznetsov, Pavel Shvechikov, Alexander Grishin, and Dmitry Vetrov.
\newblock Controlling overestimation bias with truncated mixture of continuous distributional quantile critics.
\newblock In Hal~Daumé III and Aarti Singh, editors, \emph{Proceedings of the 37th International Conference on Machine Learning}, volume 119 of \emph{Proceedings of Machine Learning Research}, pages 5556--5566. PMLR, 13--18 Jul 2020.
\newblock URL \url{https://proceedings.mlr.press/v119/kuznetsov20a.html}.

\bibitem[Liang et~al.(2018)Liang, Liaw, Nishihara, Moritz, Fox, Goldberg, Gonzalez, Jordan, and Stoica]{liang_rllib_2018}
Eric Liang, Richard Liaw, Robert Nishihara, Philipp Moritz, Roy Fox, Ken Goldberg, Joseph Gonzalez, Michael Jordan, and Ion Stoica.
\newblock {RL}lib: Abstractions for distributed reinforcement learning.
\newblock In Jennifer Dy and Andreas Krause, editors, \emph{Proceedings of the 35th International Conference on Machine Learning}, volume~80 of \emph{Proceedings of Machine Learning Research}, pages 3053--3062. PMLR, 10--15 Jul 2018.
\newblock URL \url{https://proceedings.mlr.press/v80/liang18b.html}.

\bibitem[Lillicrap et~al.(2016)Lillicrap, Hunt, Pritzel, Heess, Erez, Tassa, Silver, and Wierstra]{lillicrap_continuous_2016}
Timothy~P. Lillicrap, Jonathan~J. Hunt, Alexander Pritzel, Nicolas Heess, Tom Erez, Yuval Tassa, David Silver, and Daan Wierstra.
\newblock Continuous control with deep reinforcement learning, 2016.
\newblock URL \url{http://arxiv.org/abs/1509.02971}.
\newblock arXiv:1509.02971 [cs, stat].

\bibitem[Meier et~al.(2015)Meier, Honegger, and Pollefeys]{7140074}
Lorenz Meier, Dominik Honegger, and Marc Pollefeys.
\newblock Px4: A node-based multithreaded open source robotics framework for deeply embedded platforms.
\newblock In \emph{2015 IEEE International Conference on Robotics and Automation (ICRA)}, pages 6235--6240, 2015.
\newblock \doi{10.1109/ICRA.2015.7140074}.

\bibitem[Raffin et~al.(2021)Raffin, Hill, Gleave, Kanervisto, Ernestus, and Dormann]{ran_stable-baselines3_2021}
Antonin Raffin, Ashley Hill, Adam Gleave, Anssi Kanervisto, Maximilian Ernestus, and Noah Dormann.
\newblock Stable-baselines3: Reliable reinforcement learning implementations.
\newblock \emph{Journal of Machine Learning Research}, 22\penalty0 (268):\penalty0 1--8, 2021.
\newblock URL \url{http://jmlr.org/papers/v22/20-1364.html}.

\bibitem[Schulman et~al.(2015)Schulman, Levine, Abbeel, Jordan, and Moritz]{schulman_trust_2015}
John Schulman, Sergey Levine, Pieter Abbeel, Michael Jordan, and Philipp Moritz.
\newblock Trust region policy optimization.
\newblock In Francis Bach and David Blei, editors, \emph{Proceedings of the 32nd International Conference on Machine Learning}, volume~37 of \emph{Proceedings of Machine Learning Research}, pages 1889--1897, Lille, France, 07--09 Jul 2015. PMLR.
\newblock URL \url{https://proceedings.mlr.press/v37/schulman15.html}.

\bibitem[Schulman et~al.(2016)Schulman, Moritz, Levine, Jordan, and Abbeel]{schulman_high-dimensional_2015}
John Schulman, Philipp Moritz, Sergey Levine, Michael Jordan, and Pieter Abbeel.
\newblock High-dimensional continuous control using generalized advantage estimation.
\newblock In \emph{Proceedings of the International Conference on Learning Representations (ICLR)}, 2016.

\bibitem[Schulman et~al.(2017)Schulman, Wolski, Dhariwal, Radford, and Klimov]{schulman_proximal_2017}
John Schulman, Filip Wolski, Prafulla Dhariwal, Alec Radford, and Oleg Klimov.
\newblock Proximal {Policy} {Optimization} {Algorithms}, August 2017.
\newblock URL \url{http://arxiv.org/abs/1707.06347}.
\newblock arXiv:1707.06347 [cs].

\bibitem[Serrano-Munoz et~al.(2023)Serrano-Munoz, Chrysostomou, B{\o}gh, and Arana-Arexolaleiba]{serrano2023skrl}
Antonio Serrano-Munoz, Dimitrios Chrysostomou, Simon B{\o}gh, and Nestor Arana-Arexolaleiba.
\newblock skrl: Modular and flexible library for reinforcement learning.
\newblock \emph{Journal of Machine Learning Research}, 24\penalty0 (254):\penalty0 1--9, 2023.

\bibitem[Tian(2020)]{Tian2020Reinforcement}
Jun Tian.
\newblock Reinforcementlearning.jl: A reinforcement learning package for the julia programming language, 2020.
\newblock URL \url{https://github.com/JuliaReinforcementLearning/ReinforcementLearning.jl}.

\bibitem[Towers et~al.()Towers, Terry, Kwiatkowski, Balis, de~Cola, Deleu, Goulão, Kallinteris, KG, Krimmel, Perez-Vicente, Pierré, Schulhoff, Tai, Tan, and Younis]{Towers_Gymnasium}
Mark Towers, Jordan~K Terry, Ariel Kwiatkowski, John~U. Balis, Gianluca de~Cola, Tristan Deleu, Manuel Goulão, Andreas Kallinteris, Arjun KG, Markus Krimmel, Rodrigo Perez-Vicente, Andrea Pierré, Sander Schulhoff, Jun~Jet Tai, Andrew Jin~Shen Tan, and Omar~G. Younis.
\newblock {Gymnasium}.
\newblock URL \url{https://github.com/Farama-Foundation/Gymnasium}.

\bibitem[Verschueren et~al.(2022)Verschueren, Frison, Kouzoupis, Frey, Duijkeren, Zanelli, Novoselnik, Albin, Quirynen, and Diehl]{verschueren_acadosmodular_2022}
Robin Verschueren, Gianluca Frison, Dimitris Kouzoupis, Jonathan Frey, Niels~Van Duijkeren, Andrea Zanelli, Branimir Novoselnik, Thivaharan Albin, Rien Quirynen, and Moritz Diehl.
\newblock acados—a modular open-source framework for fast embedded optimal control.
\newblock \emph{Mathematical Programming Computation}, 14\penalty0 (1):\penalty0 147--183, March 2022.
\newblock ISSN 1867-2949, 1867-2957.
\newblock \doi{10.1007/s12532-021-00208-8}.
\newblock URL \url{https://link.springer.com/10.1007/s12532-021-00208-8}.

\bibitem[Weng et~al.(2022)Weng, Chen, Yan, You, Duburcq, Zhang, Su, Su, and Zhu]{weng_tianshou_2022}
Jiayi Weng, Huayu Chen, Dong Yan, Kaichao You, Alexis Duburcq, Minghao Zhang, Yi~Su, Hang Su, and Jun Zhu.
\newblock Tianshou: A highly modularized deep reinforcement learning library.
\newblock \emph{Journal of Machine Learning Research}, 23\penalty0 (267):\penalty0 1--6, 2022.
\newblock URL \url{http://jmlr.org/papers/v23/21-1127.html}.

\end{thebibliography}

\end{document}